\documentclass[a4paper,alpha-refs]{oup-contemporary}

\journal{gigascience}

\usepackage{graphicx}
\usepackage{siunitx}

\usepackage{xcolor}
\usepackage{makecell}
\usepackage{tabularx}
\usepackage{makecell}
\usepackage{adjustbox}
\usepackage{caption}
\usepackage{subcaption}
\usepackage{nicefrac}
\usepackage{float}
\usepackage{mathtools}
\usepackage[textsize=tiny]{todonotes}
\usepackage{cancel}
\usepackage{orcidlink}

\newcolumntype{R}[2]{%
    >{\adjustbox{angle=#1,lap=\width-(#2)}\bgroup}%
    l%
    <{\egroup}%
}
\newcommand*\rot{\multicolumn{1}{R{90}{0.5em}}}

\newcommand{\rotsmash}[1]{\quad\smash{\raisebox{-5.5em}{\rotatebox{90}{#1}}}}

\newcommand{\bra}[1]{\left[ #1 \right]}
\newcommand{\ie}{\emph{i.e.}}
\newcommand{\eg}{\emph{e.g.}}

\title{Benchmarking missing-values approaches for predictive models on
health databases.}

\author[1,2,3,\authfn{1}]{Alexandre~Perez-Lebel~\orcidlink{0000-0003-0556-0763}}
\author[1,2,3]{Gaël~Varoquaux~\orcidlink{0000-0003-1076-5122}}
\author[2]{Marine~Le~Morvan~\orcidlink{0000-0001-9899-221X}}
\author[4,5]{Julie~Josse~\orcidlink{0000-0001-9547-891X}}
\author[1]{Jean-Baptiste~Poline~\orcidlink{0000-0002-9794-749X}}

\affil[1]{McConnell Brain Imaging Centre, The Neuro (Montreal Neurological Institute-Hospital), Faculty of Medicine, McGill University, Montreal, QC, Canada}
\affil[2]{Inria, Parietal team, Palaiseau, France}
\affil[3]{Mila, Montreal, QC, Canada}
\affil[4]{Inria, Montpellier, France}
\affil[5]{IDESP, Montpellier, France}

\authnote{\authfn{1}\href{mailto:alexandre.perez@inria.fr}{alexandre.perez@inria.fr}
\orcidlink{0000-0003-0556-0763}
 }

\doi{Published version DOI: 10.1093/gigascience/giac013}
\papercat{Research}
\manuscripttype{Author Accepted Manuscript}
\footline{This is the author accepted manuscript. The published version is available at DOI: 10.1093/gigascience/giac013.}

\runningauthor{Perez-Lebel et al.}

\begin{document}

\begin{frontmatter}
    \maketitle
    \begin{abstract}
        \textbf{BACKGROUND}
        As databases grow larger, it becomes harder to fully control their
        collection, and they frequently come with missing
        values: incomplete observations. These large databases are well suited
        to train machine-learning models, for instance for forecasting or
        to extract biomarkers in biomedical settings. Such predictive approaches can
        use discriminative --rather than generative-- modeling, and thus open the
        door to new missing-values strategies. Yet existing empirical evaluations
        of strategies to handle missing values have focused on inferential
        statistics.
        \textbf{RESULTS}
        Here we conduct a systematic benchmark of missing-values strategies in
        predictive models with a focus on large health databases: four electronic health
        record datasets, a population brain imaging one, a health survey and two intensive care ones.
        Using gradient-boosted trees, we compare native support for missing
        values with simple and state-of-the-art imputation prior to learning.
        We investigate prediction accuracy and computational time. For
	prediction after
        imputation, we find that adding an indicator to express which values have been imputed is important, suggesting that the data are missing not at random.
        Elaborate missing values imputation can improve prediction
	compared to
        simple strategies but requires longer computational time on large
        data. Learning trees that model
        missing values --with missing incorporated attribute-- leads to
	robust, fast, and well-performing predictive modeling.
        \textbf{CONCLUSIONS}
        Native support for missing values in supervised machine learning
        predicts better than state-of-the-art imputation with
	much less
        computational cost. When using imputation, it is important to add
        indicator columns expressing which values have been imputed.
    \end{abstract}

\end{frontmatter}

\begin{keypoints*}
    \begin{itemize}
        \item Benchmarks on health databases highlight the challenges
	that they represent for statistical learning:
	non-ignorable missing values (Missing Not At Random -- MNAR),
	non-linear relationships between covariates and outcomes.
        \item Native missing-values support in supervised machine learning gives better prediction than state-of-the-art imputation with significantly less computational cost.
        \item With linear models, conditional imputation is to be preferred.
        \item When using imputation, concatenating the missingness indicator with the input features significantly improves predictions.
        \item Bagging, as sometimes used for multiple-imputation, improves prediction performance but with a prohibitive time cost.

    \end{itemize}
\end{keypoints*}

\section{Background: missing values in databases}
\label{sec:background}

Missing values are pervasive in many application domains. This is
particularly true on health data, where missing values arise for a
multitude of reasons: two patients rarely follow the same medical path
and take the exact same set of exams; measurements are
omitted because of lack of time or because the patient's
condition does not allow it; hospitals do not collect exactly the same
information because of diverging practices and the use of different
devices; etc. This problem is exacerbated when the data are aggregated
across multiple sources or when each individual
sample comprises many features. The more data there is, the more data is missing.


There is a rich and established statistical literature for the treatment
of missing data \citep{little2019statistical,wells2013strategies},
which has so far been
mostly focused on inferential purposes, \ie{} estimating parameters of a
probabilistic model with their confidence intervals. For such problem, an
important distinction between missing data mechanisms was introduced by
\citet{Rubin1976}: Missing Completely At Random (MCAR) where the
probability of having missing data does not depend on the covariates,
Missing At Random (MAR), where the probability of a missing value only
depends on the observed values of other variables; and Missing Not At
Random (MNAR) which covers all other cases. MNAR corresponds to cases
where the missingness carries information. For example, if heartbeat
measures are not reported when the values are too low, it creates a MNAR
situation. Most available methods for inference in the presence of
missing values are only valid under the MAR assumption, including maximum likelihood approaches with the Expectation Maximization algorithm \citep{Dempster1977}, as well as Multiple Imputation \citep{van2018flexible}. The latter is a two-step approach where the data is first imputed multiple times to create multiple completed datasets, and then the analysis is performed on each imputed dataset separately before combining the results to take into account the uncertainty due to missing values.

\smallskip



Supervised learning to build models that predict best a response using
covariates with missing values can lead to different tradeoff than
inference models \citep{Sperrin2020,josse2019consistency}.
In health, such predictive models
are central to building complex biomarkers or risk scores, to
forecasting an epidemic, and they can even underlie causal inference for policy
evaluation \citep{rose2020machine}.
They are increasingly used on electronic health records
\citep{miotto2016deep,zheng2017machine,steele2018machine}, where the
choice of strategy to handle missing values remains a challenge
\citep{jarrett2021clairvoyance}.
Indeed, unlike with inference, little work to date has focused on the
systematic evaluation of supervised learning with missing values.
Existing works focus on benchmarking imputation quality
\citep{Jager2021,Bertsimas2018} – which, as our study points out, is a
different goal than prediction quality – or only focus on
imputation-based methods \citep{Poulos2018}. 

In practice, a number of options are commonly used to learn predictive
models with missing values.
The simplest one is to delete all observations containing missing values. However, leaving aside the possible biases that this practice may induce, it often leads to considerable loss of information in high and even moderate dimensions. Indeed, when there are many variables, it is common that only a few observations are completely observed.


In order to deal with arbitrary subsets of input features, the most
common practice currently consists in first imputing the missing values,
and then learning a predictive model (e.g regression or classification)
on the completed data. The popularity of this approach is mainly due to
its simplicity and ease of implementation. After imputation,
off-the-shelf learners can be applied on the completed dataset. Recent theoretical results show that applying a supervised-learning regression on imputed data can asymptotically recover the optimal prediction function; however most imputation strategies, including the common imputation by the conditional expectation, create discontinuities in the regression function to learn \citep{Morvan2021}.

A small number of machine learning models can natively handle missing
values, in particular popular tree-based methods. Trees greedily
partition the input space into subspaces in order to minimize a risk.
This non-smooth optimization scheme enables them to be easily adapted to
directly learn from incomplete data. Several adaptations of trees to
missing values have been proposed \citep[see][ for a
short review]{josse2019consistency}. Missing Incorporated in Attributes
\citep[MIA,][]{Twala2008methods} is the most promising strategy
\citep{josse2019consistency}, described below in the experiment section.
\medskip

In this work, we benchmark the most popular methods for supervised
learning with missing values on multiple large real-world health databases. In
contrast to most simulations, real health databases combine a number of
challenges: unknown data distributions (not necessarily Gaussian),
uncontrolled missing data mechanism (not necessarily MAR), mixed
quantitative and categorical data, and often a high level of noise. In
such a challenging setting, we compare existing approaches to make
recommendations that are directly relevant for the practitioner. To
establish general recommendations, we study a total of 13 prediction
\emph{real-world} tasks (10 classification and 3 regression tasks) across four
publicly-available health databases of very different nature. For each of
these tasks, we compare methodologies based on imputation followed by
regression or classification, to tree-based models that can natively
handle missing values with a MIA strategy. These methods are chosen from
the common practice as well as theoretical work on supervised learning with missing values \citep{josse2019consistency}.
\medskip

The present study has several strengths in terms of benchmarking
methodology, avoiding common limitations. It uses real data \textit{and}
real missingness; multiple draws of a cross-validation loops are used;
the imputation procedure is not fitted on the whole dataset but rather on
the training set to prevent leaks from the training set to the
out-of-sample test set; hyper-parameters of the predictive model are
tuned for each method to reduce bias in the hyper-parameters selection;
and finally the study benchmarks imputation methods \textit{and} predictive models that handles missing values.
As a result, our benchmark is very computation-intensive: the whole study
costed approximately 520\,000 CPU hours, \ie{} 60 years on a single CPU,
revealing the need to also account for compute cost in recommendations.
\medskip

After briefly exposing our benchmarking methodology, we give a synthetic
view of the findings and discuss observed trends. Overall, the benchmarks
reveal the presence of missing not at random (MNAR) values and non-linear
mechanisms. High-quality conditional imputation gives good prediction
provided that a variable indicating which entries were imputed is
added to the completed data. However, its algorithmic complexity makes
it prohibitively costly on large data. Rather, tree-based methods with
integrated support for missing values (missing incorporated attribute --
MIA) perform as well or better, at a fraction of the computational cost.

\section{Empirical study}

\subsection{Benchmarking the imputation and MIA methods}


Our experiments compare two-step procedures based on imputation followed by regression or classification, as well as tree-based models with an intrinsic support for missing values thanks to MIA.
The 12 methods compared are summarized in Table~\ref{tab:methods:tree}:
MIA, 8 methods based on single imputation and 3 methods using Multiple
Imputation via Bagging. Below, we describe further the imputation strategies benchmarked as well as MIA.

\begin{table}[b!]
    \caption{\textbf{Methods compared in the main experiment.}\\
    All use gradient-boosted trees as predictive model. 10 use imputation and 2 uses MIA. Bagging uses 100 estimators in the ensemble.}
    \label{tab:methods:tree}
    \setlength\tabcolsep{4.2pt}
    \begin{tabularx}{\linewidth}{l l l l l}
    \toprule
    \makecell{In-article name} & \makecell{Imputer} & \makecell{Mask} & \makecell{Bagging} & \makecell{Predictive model} \\
    \midrule
    MIA & – & No & No & Boosted trees \\
    Mean & Mean & No & No & Boosted trees \\
    Mean+mask & Mean & Yes & No & Boosted trees \\
    Median & Median & No & No & Boosted trees \\
    Median+mask & Median & Yes & No & Boosted trees \\
    Iterative & Iterative & No & No & Boosted trees \\
    Iterative+mask & Iterative & Yes & No & Boosted trees \\
    KNN & KNN & No & No & Boosted trees \\
    KNN+mask & KNN & Yes & No & Boosted trees \\
    \hline
    \makecell[l]{Iterative\vspace{-0.1cm}\\+Bagging} & Iterative & No & Yes (100) & Boosted trees \\
    \makecell[l]{Iterative+mask\vspace{-0.1cm}\\+Bagging} & Iterative & Yes & Yes (100) & Boosted trees \\
    MIA+Bagging & – & No & Yes (100) & Boosted trees \\
    \bottomrule
\end{tabularx}

\end{table}

\subsubsection{Single Imputation}


\paragraph{Constant imputation: mean and median}

The simplest approach to imputation is to replace missing values by a constant such as
the mean, the median or the mode of the corresponding feature. This is frowned upon in
classical statistical practice, as the resulting data distribution is severely
distorted compared to that of fully-observed data. Yet, in a supervised setting, the goal is different from that of inferential tasks. Recent theoretical results have established that
powerful learners such as ones based on trees can learn to recognize such
imputed values and give the best possible predictions
\citep{josse2019consistency}. The key to the success of this strategy is to impute the training and the test set with the same constant: missing values of the test set are imputed with the constants learned on the training set (mean, median, etc).

\paragraph{Conditional imputation: MICE and KNN}

Powerful imputation approaches rely on conditional dependencies between
features to fill in the missing values. Adapting machine-learning
techniques gives flexible estimators of these dependencies. Classical
approaches include k-nearest neighbor regressors \citep{chen2000nearest},
and iterative conditional imputers that predict one feature as a function
of others, as with the MICE imputer \citep{buuren2010mice}. In our
experiments, we benchmark their implementation in scikit-learn
\citep{pedregosa2011scikit}: the {\tt KNNImputer} as well as the {\tt
IterativeImputer}, using
linear models to impute missing values.

\paragraph{Adding the mask}

Conditional imputation can make it hard for the learner to retrieve which entries were originally observed and which were originally missing. However, the information of missingness can be relevant for predicting the outcome in cases where it depends on missingness, or in \emph{missing not at random} settings where the missingness carries information. For these reasons, it can be useful after imputation to add new binary features that encode whether a value was originally missing or not: the \emph{mask} or
\emph{missingness indicator} \citep{josse2019consistency,
Sharafoddini2019,Sperrin2020}.

\subsubsection{Multiple Imputation}

When estimating model parameters, it is of great importance to
reflect the uncertainty due to the missing values. For this purpose,
Multiple Imputation methods are widely used, often via Resampling methods
such as the Bootstrap. However, for prediction (classification or
regression) theoretical conditions differ from that of parameters estimation.
Indeed, it has been shown recently that a sufficiently flexible learner
reaches optimal performances asymptotically with Single Imputation, whatever
the missing data mechanism and whatever the choice of imputation function
\citep{Morvan2021}. Still, this result holds in asymptotic regimes, and
there is a need for empirical results on handling missing values with
Multiple Imputation or Bootstrap in the context of supervised learning.
Theoretically, the only result that we are aware of for Multiple Imputation in the context of prediction requires access to an oracle predictor for fully observed data and is valid only in MAR \citep[th.\,3]{josse2019consistency}.
In general, it is not clear how to use Multiple Imputation
for supervised learning: sampling can be applied in different ways during
training the model or applying their predictions to new data.
\citet{khan2019bootstrapping} review and compare a number of methods for
using Multiple Imputation and Bootstrap: learning on an averaged version
of a multiply imputed dataset, bagging single imputations, bagging
Multiple Imputations, constructing ensembles based on predictors that
were each learned on a version of a multiply imputed dataset  \citep[chap
16]{friedman2001elements}. As these methods all come with a significant
computing cost, we focus on the most promising approach: bagging single
imputation. More precisely, we draw for each task 100 bootstrap
replicates. We then fit the single imputation and the predictive model
on each of these replicates, to obtain 100 predictors. Final predictions are made either by voting or by averaging (see Supplementary Table~\ref{tab:methods:sklearn}).

\subsubsection{Directly handling missing values with tree-based models: MIA}
We also consider the MIA (Missing Incorporated in Attribute) strategy to
readily model missing values in tree-based models. It has the benefit
of using all samples, including incomplete ones, to produce the splits of
the input space. More precisely for each split based on variable $j$, all
samples with a missing value in variable $j$ are either sent to the left
or to the right child node depending on which option leads to the lowest
risk. Note that the samples with an observed value in variable $j$ can
either be split between the left and right child node according to
whether their values $x_j$ is greater or smaller than a threshold, or
either all be sent to the same child node so that they are separated from
the samples with a missing value in variable $j$. That makes MIA
particularly suited to Missing Not At Random (MNAR) settings, as it can
harness the missingness information. Moreover, since trees with MIA
directly learn with missing values, they provide a straightforward way of
dealing with missing values in the test set. We use the implementation in
scikit-learn's boosted trees ({\tt
HistGradientBoostingRegressor}).

\subsubsection{Predictive model}
For the supervised learning step, we focus on gradient-boosted trees
--though we also benchmark linear models in a complementary analysis
described in the appendices. We applied supervised learning to the imputed
data for the imputation-based methods. We also used the tree models with
their support of MIA for a direct handling of missing values. Gradient-boosted trees are state-of-the art predictors for tabular data
\citep{chen2016xgboost,olson2018data,shwartz2021tabular}, and thus constitute a strong
baseline. Moreover, using gradient-boosted trees enables us to keep the same predictive model for all approaches, thereby putting emphasis on the impact of the missing data treatment.

To define the input features we either use the choice of experts in prior
studies, or feature screening, a classic machine-learning procedure using
a simple ANOVA-based univariate test of the link of each feature to the
outcome
\citep{guyon2003introduction}.
In both cases, the same set of selected features is used for all methods within each predictive task.
Selecting features is necessary because some of the
imputation methods studied are not tractable with a large number of
features.




\subsection{Health databases}
\label{sec:datadescription}
To reach conclusions as general as possible we used four real-world health-related databases. These databases vary in terms of location, size, purpose
and time, to cover a wider data scope. These databases already existed
and no data collection was made in this study. Below, we describe them
briefly, giving the prediction tasks studied for each of them.

\subsubsection{Traumabase}
\cite{Traumabase} is a collaboration studying major trauma. The database gathers information from 20 French trauma centers on more than 20\,000 trauma cases from admission until discharge from critical care. Data collection started in 2010 and is still ongoing in 2020. We used records spanning from 2010 to 2019. Data can be obtained by contacting the team on the Traumabase website \citep{Traumabase}.

We defined 5 prediction tasks on this database, 4 classifications and 1 regression. Outcomes are diverse: patient's death, hemorrhagic shock, septic shock, and platelet count. Features for the hemorrhagic shock prediction are taken from \citet{Jiang2020a}.

\subsubsection{UK Biobank}
UK Biobank (UKBB) \citep{Sudlow2015} is a major prospective epidemiology cohort with biomedical measurements. It provides health information on more than 500\,000 United-Kingdom participants aged between 40 to 69 years from 2006 to 2010. The data are available upon application as detailed on the UK BioBank website \citep{Sudlow2015}.

We defined 5 tasks on this database, 4 classifications and 1 regression.
Outcomes are the diagnosis of three diseases - breast cancer, skin
cancer, Parkinson's disease - as well as prediction of the
fluid-intelligence score. Breast cancer prediction uses features defined in \citet{Lall2019}.

\subsubsection{MIMIC-III}
The Medical Information Mart for Intensive Care (MIMIC) database
\citep{Johnson2016} is an Intensive Care Unit (ICU) dataset developed by
the MIT Lab for Computational Physiology. It comprises deidentified
health data associated with about 60\,000 ICU admissions recorded at the Beth Israel Deaconess Medical Center of Boston, United States, between 2001 and 2012. It includes demographics, vital signs, laboratory tests, medications, and more. The data can be accessed via an application described on the MIMIC website \citep{Johnson2016}.

We defined 2 classification tasks on this database. Outcomes are septic shock and hemorrhagic shock.

\subsubsection{NHIS}
The National Health Interview Survey (NHIS) \citep{NHIS2017} is a major data collection program of the National Center for Health Statistics (NCHS), part of the Centers for Disease Control and Prevention (CDC) in the United States. It aims to monitor the health of the population. Since 1957, it collects data from United-States population. We use the 2017 edition, summing up to approximately 35\,000 households containing about 87\,500 persons. The database is freely-accessible  on the NHIS website \citep{NHIS2017}.

We defined 1 regression task on this database. Outcome is the yearly income.
\smallskip

More details on each database and task can be found in the appendices, in
particular in Supplementary Table~\ref{tab:tasks} and Supplementary
Figure~\ref{fig:data:types} that detail the number of features available
and their type (numerical, ordinal and categorical), and Supplementary
Figure~\ref{fig:data:mv} giving the distribution of missing values across features.

\begin{figure*}[t!]
    \begin{subfigure}{\linewidth}
        \centering
        \subcaption{{\normalsize\textbf{Prediction performance}}}
        \label{fig:finding:trees:score}
        \includegraphics[width=\textwidth]{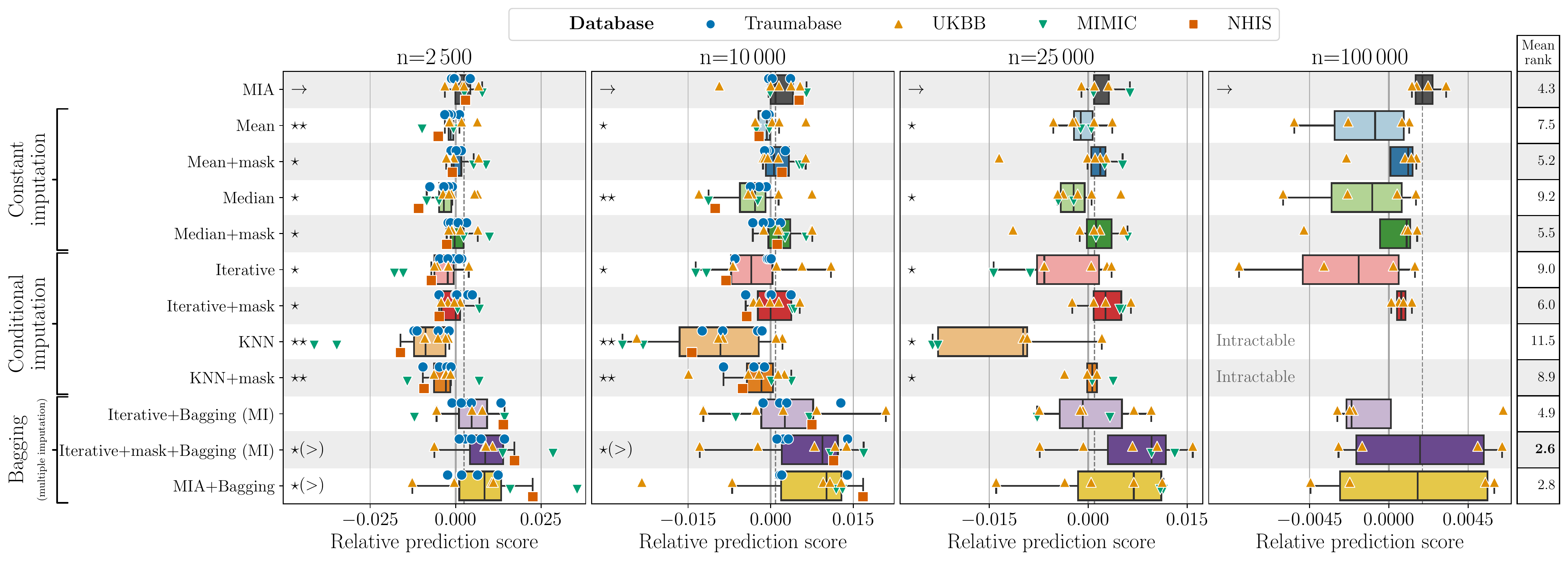}
    \end{subfigure}

    \medskip
    \begin{subfigure}{\linewidth}
        \centering
        \subcaption{{\normalsize\textbf{Computational time}}}
        \label{fig:finding:trees:time}
        \includegraphics[width=\textwidth]{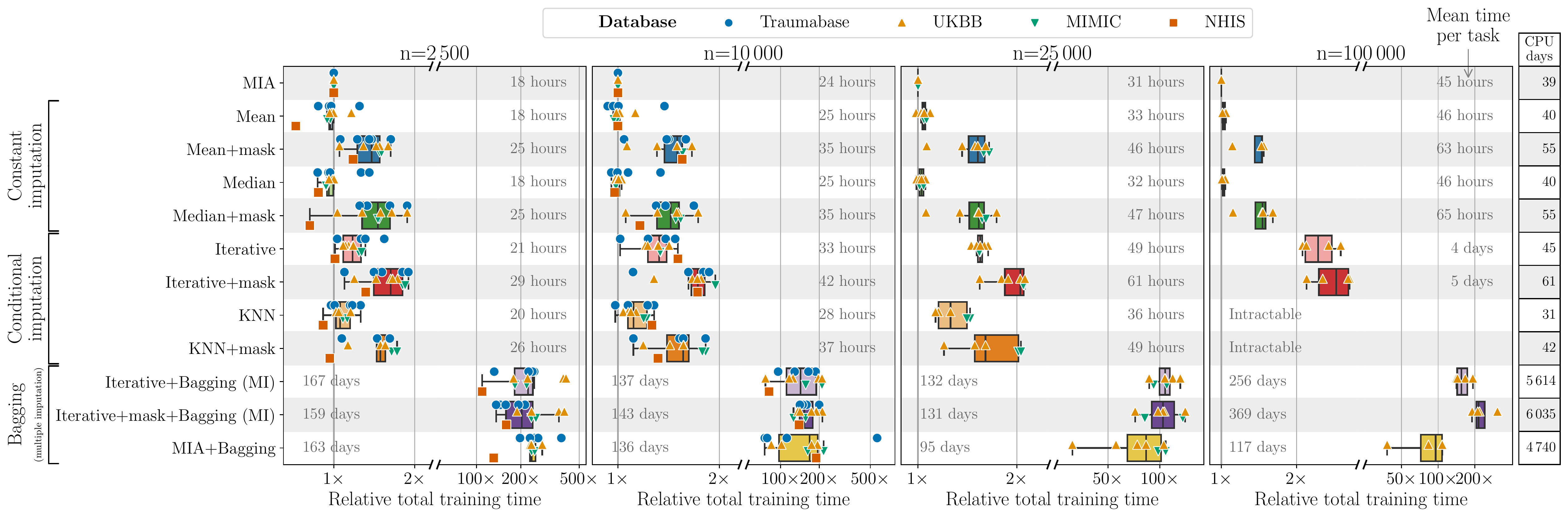}
    \end{subfigure}
    \caption{\textbf{Gradient-boosted trees models.}
    Comparison of prediction performance and training times across the 12 methods (see Table~\ref{tab:methods:tree}) for 13 prediction tasks spread over 4 databases, and for 4 sizes of dataset (2\,500, 10\,000, 25\,000 and 100\,000 samples). For each of the tasks and sizes, we computed a reference score by averaging the scores obtained by the 12 methods on the corresponding task and size. The relative prediction score of a method on a task and size is the deviation of the prediction score from the reference score of this task and size. For computational time, the total training time comprises imputation and tuning times and is given relative to the one of MIA for each task and size. More details on how these plots were created are given in the \textit{\nameref{sec:plotting-method}} section. The significance is assessed with a one-sided Wilcoxon signed-rank test with MIA taken as reference (see Supplementary Table~\ref{tab:wilcoxon:mia-vs-rest}). Methods which performed significantly poorer (resp. better) at the 0.05 level are marked with "$\star$" (resp. "$\star(>)$") and "$\star\star$" (resp. "$\star\star(>)$") for Bonferroni-corrected levels. Two tables give the overall average ranks and the total number of CPU days for each method, all tasks and sizes combined. The average number of CPU hours \textit{per task} required to evaluate each method is given on each line. Detailed scores and ranks broken out by tasks are given in Supplementary Table~\ref{tab:tree:scores-ranks} and Supplementary Figure~\ref{fig:data:breakout}. Notice that KNN and KNN+mask were intractable at $n=100\,000$ due to their memory footprint of $\mathcal{O}(n^2)$.}
    \label{fig:finding:trees}
\end{figure*}


\subsection{Findings}

Figure~\ref{fig:finding:trees} summarizes the performances and
computational times of the various methods across the 4 databases and 13
prediction tasks. To explore the importance of the amount of data, we
created training datasets of 4 sizes: 2\,500, 10\,000, 25\,000 and
100\,000 samples. We report the general trends.

\subsubsection{Bagging improves prediction, MIA performs well at limited
cost}
Iterative+mask+Bagging obtains the best overall average rank (2.6) across all tasks and
sizes in terms of prediction score closely followed by MIA+Bagging (2.8)
as shown on Figure~\ref{fig:finding:trees:score} and Supplementary
Table~\ref{tab:tree:ranks}. Overall, Bagging improves markedly all
approaches (supplementary Figure~\ref{fig:finding:bagging}).
However the cost of these bagged methods can be prohibitive. At size $n$=100\,000, Iterative+mask+Bagging and MIA+Bagging cost 369 and 117 CPU days per task respectively, about 100 to 200 times slower than non-bagged method such as MIA (1.9 CPU days per task).

MIA enables to navigate a trade off between prediction performance and computational
tractability: With Bagging it comes close to Iterative+Mask with half the
computational cost on large databases. Without Bagging, it is the best overall
performer, with an overall average rank of 4.3, and up to 200 times
faster. It is
followed by Mean+mask, Median+mask and Iterative+mask with overall average ranks of 5.2, 5.5 and 6.0 respectively. Mean, KNN+mask, Iterative, Median and KNN performed the worst with overall average ranks of 7.5, 8.9, 9.0, 9.2 and 11.5 respectively. Supplementary Table~\ref{tab:tree:scores}~and~\ref{tab:tree:ranks}
give more quantitative details about scores and ranks of each method.

\begin{figure}[H]
    \centering
    \caption{\textbf{Mean ranks by method and by size of dataset.} The critical difference is computed using the Nemenyi test (equation~\eqref{eq:test:cd} and Supplementary Table~\ref{tab:friedman}). Methods within the critical difference range do not perform significantly differently from one another according to the Nemenyi test. Methods within the critical difference range of MIA are in red, others in black.}

    \begin{subfigure}{0.49\textwidth}
        \centering
        \subcaption{Tree-based methods.}
        \includegraphics[width=\linewidth]{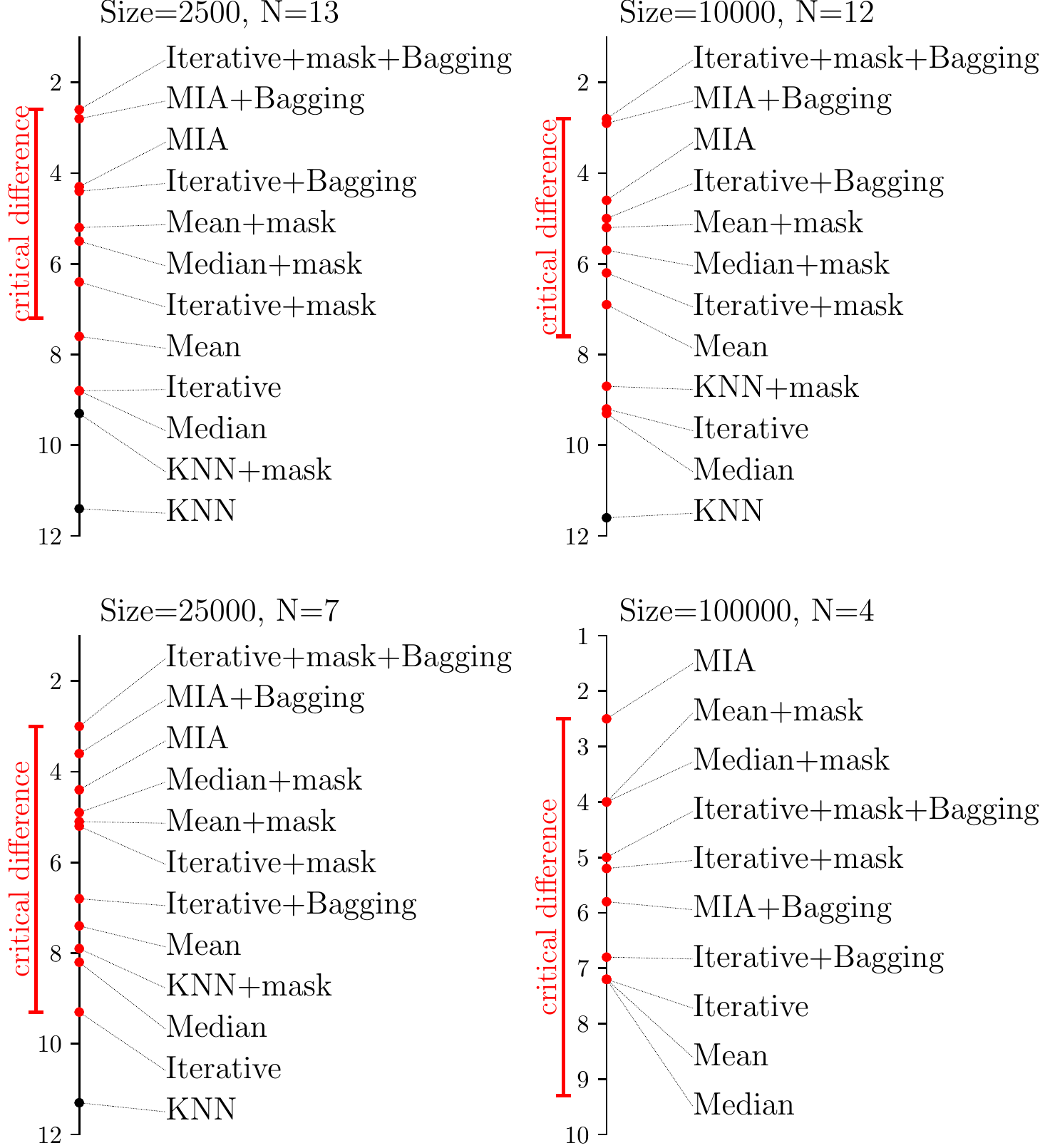}
        \label{fig:trees:nemenyi}
    \end{subfigure}
    \begin{subfigure}{0.49\textwidth}
        \centering
        \subcaption{Boosted trees+MIA vs linear methods.}
        \includegraphics[width=\linewidth]{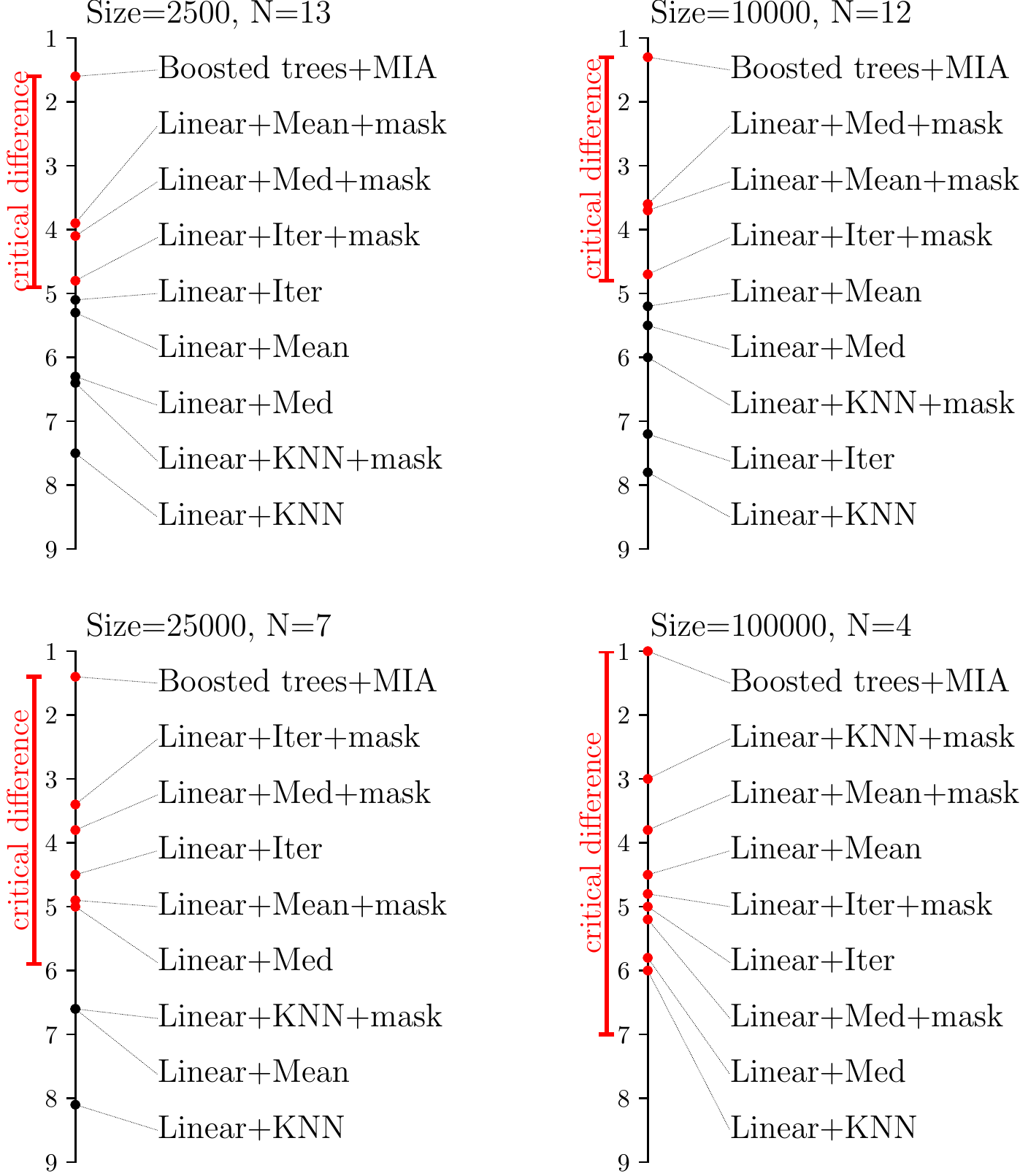}
        \label{fig:linear:nemenyi}
    \end{subfigure}

    \label{fig:my_label}
\end{figure}

Similar observations can be made on each size separately.
MIA obtained the best prediction scores on every size with average ranks of 4.3, 4.6, 4.4 and 2.5 on sizes 2\,500, 10\,000, 25\,000 and 100\,000 respectively as shown on Supplementary Figure~\ref{fig:trees:nemenyi}.


In terms of computing time, beyond the fact that Bagging multiplies by 100
the cost of every method, MIA is almost always the fastest
(Figure~\ref{fig:finding:trees:time}), though it gives excellent
prediction performance. It is on par with Mean, and Median imputations,
but adding the mask to these methods --a key ingredient to prediction
performance-- doubles their computing times. At the other end of the spectrum, Iterative+mask and KNN+mask are the slowest non-bagged methods. The gaps between training times of the methods increase with the size of the database, revealing the difference in algorithmic scalability.

\paragraph{Statistical significance}
To assess significance of the above results, we ran three statistical tests: the Friedman test \citep{Friedman1937, Friedman1940}, the Nemenyi test \citep{nemenyi1963distribution} and the one-sided Wilcoxon signed-rank test \citep{Wilcoxon1945}, all described in \citet{Demsar2006}.\\
The Friedman test compares the average ranks of several algorithms ran on several datasets. The null hypothesis assumes that all algorithms are equivalent, \ie{} their rank should be equal. Table~\ref{tab:trees:friedman} shows that the null hypothesis is rejected with p-values way below the 0.05 level for the sizes 2\,500, 10\,000 and 25\,000. This indicates that at least one algorithm has significantly different performances from one other on these sizes.
Following \citet{Demsar2006}, we then proceed with a post-hoc analysis with the Nemenyi test, assessing the significance of the difference between two algorithms using a critical difference. Algorithms with a difference in ranks smaller than the critical difference are not significantly different. Unfortunately, there are many methods to compare (12) comparatively to the number of datasets (13). As a result, the critical difference is high as shown in equation \eqref{eq:test:cd} and Supplementary Table \ref{tab:trees:friedman} and there is almost no significance when comparing the performance of MIA with the one of the other methods as shown on Supplementary Figure~\ref{fig:trees:nemenyi}. However, there are some significance when comparing bagged methods. For example at size $n$=2\,500, Iterative+mask+Bagging and MIA+Bagging performed significantly better than Mean, Median, Iterative, KNN+mask and KNN.

We run a complementary analysis with a one-sided Wilcoxon signed-rank test, used for non-parametric tests comparing algorithms pairwise. We compare MIA to every other methods. The null hypothesis claims that the median of the score differences between the two methods is positive (resp. negative) for the one-sided right (resp. one-sided left) test. Results of the test is shown on Figure~\ref{fig:finding:trees:score} and Supplementary Table~\ref{tab:wilcoxon:gbt-vs-linear}. At size $n$=2\,500, MIA performed significantly better than every other non-bagged methods at the 0.05 level. MIA also performed significantly better than Mean, KNN and KNN+mask at the Bonferroni-corrected level. Bagged methods Iterative+mask+Bagging and MIA+Bagging performed significantly better than MIA at the 0.05 level. The bigger the size $n$, the less tasks are available and so the less significant are the results.

\subsubsection{Adding the mask improves prediction}

Imputations with the additional variable representing the mask perform systematically better in terms of average prediction score than their counterpart without mask (Figure \ref{fig:finding:trees:score}, Supplementary Table~\ref{tab:tree:ranks}).

In addition, MIA is not significantly better than the masked imputations yet it is for the non-masked imputations (Figure~\ref{fig:finding:trees:score}, Supplementary Table~\ref{tab:wilcoxon}). However, adding the mask leads to longer training times (Figure~\ref{fig:finding:trees:time}). Indeed, adding the mask doubles the number of features for the supervised-learning step.

\subsubsection{Conditional imputation is on par with constant imputation}
Figure~\ref{fig:finding:trees} shows that conditional imputation using Iterative or KNN imputers does not perform consistently better than constant imputation. The overall mean rank of Iterative and KNN are 9.0 and 11.5 versus 7.5 and 9.2 for Mean and Median respectively (Figure~\ref{fig:finding:trees:score} and Supplementary Table~\ref{tab:tree:ranks}), and a similar delta is visible on the masked version.

\subsubsection{Supplementary finding: Boosted-trees outperform linear methods}
Imputation methods paired with a linear model performed poorer than when
paired with boosted-trees (Supplementary Figure~\ref{fig:finding:linear}, Supplementary Table
\ref{tab:linear:ranks}). Additionally, boosted-trees paired with MIA
are significantly better than every other method based on a linear model (Supplementary Table
\ref{tab:wilcoxon}).

\section{Discussion}

\subsection{Interpretation}

\subsubsection{Model aggregation drives the good performance of
Multiple Imputation}

As with standard multiple-imputation strategies used for parameter
estimation, Bagging generates multiple bootstrap replicate training sets.
Yet, the standard practice of Multiple Imputation strives to capture well
the conditional distribution of the missing values given the observed
one, while such conditional imputation is not needed for good prediction
\citep[as revealed by the good performance of MIA and ][]{Morvan2021}.
Indeed, Bagging in itself is known to improve generalization. To answer
whether the good performance of Multiple Imputation can be attributed to
ensembling --averaging multiple predictors-- or capturing the conditional
distribution, we performed additional experiment with Mean+mask+Bagging
(see Supplementary Figure~\ref{fig:finding:bagging}). We observed that
Mean+mask+Bagging is on par with Iterative+mask+Bagging, which suggests that the improved performances are rather due to the effect of Bagging itself rather than
capturing the conditional distribution of the missing data given the observed ones.

\subsubsection{Good imputation does not imply good prediction, even for Multiple Imputation}
It may be surprising at first that a sophisticated conditional imputation
does not outperform constant imputation. Indeed, it contradicts the
intuition that better imputation should lead to better prediction.
Theoretical work shows that this intuition is not always true
\citep{Morvan2021}: even in MAR settings, it may not hold for strongly non-linear
mechanisms and little dependency across features.
In the health databases that we studied, the features are weakly
correlated: on average, only 12\% of the features
are correlated at more than 0.3 in absolute value
(Supplementary Table \ref{tab:data:cor}). This low correlation among
features may explain our findings.
If features are mostly independent,
there is little information on
the unobserved values to be extracted from the observed ones.
For supervised learning, constant imputation comes with the benefit that
it creates a simple structure captured by the supervised-learning step,
which can then adapt to the missingness
\citep{josse2019consistency}.

\subsubsection{Boosted-trees with MIA give best predictive models at
little cost}

MIA, the missing-values support inside gradient-boosted trees, appears as
a method of choice to deal with missing values. It was on average the
best performing one in terms of performance in our extensive benchmark
while having a low computational cost. Sophisticated conditional
imputation such as the Iterative or KNN imputers are appealing because
they may recover plausible values for missing entries, as discussed
below. However, they are intractable with large datasets. Beyond the
costs outlined by our experiments (Figure~\ref{fig:finding:trees:time}), the
broader problem is the algorithmic scalability: for a dataset of $p$
features and $n$ samples, the compute cost of a KNN imputer scales as
$n^2 p^2$ and the memory footprint as $n^2$, while the compute cost of an
iterator imputer scales as $p^2 n \min(n, p)$
when it is based on linear models, the cheapest alternative. If both $p$
and $n$ grow, these costs rapidly becomes prohibitive. They prevented us
from exploring larger datasets, \eg{} with more features.
Note that to ground valid predictions, the imputation model must be
learned only on the train set; hence it is recomputed many times in a
cross-validation loop.

\begin{figure}[H]
    \centering
    \caption{\textbf{Types of features.} Number of categorical, ordinal and numerical features in each dataset, \textit{before encoding}. Note that one non-encoded categorical feature can lead to several selected encoded features. Since we select 100 encoded features, some task have less than 100 non-encoded features. For tasks having several trials, five horizontal bars are plotted representing one trial each, as feature selection may select different features.}
    \includegraphics[width=\linewidth]{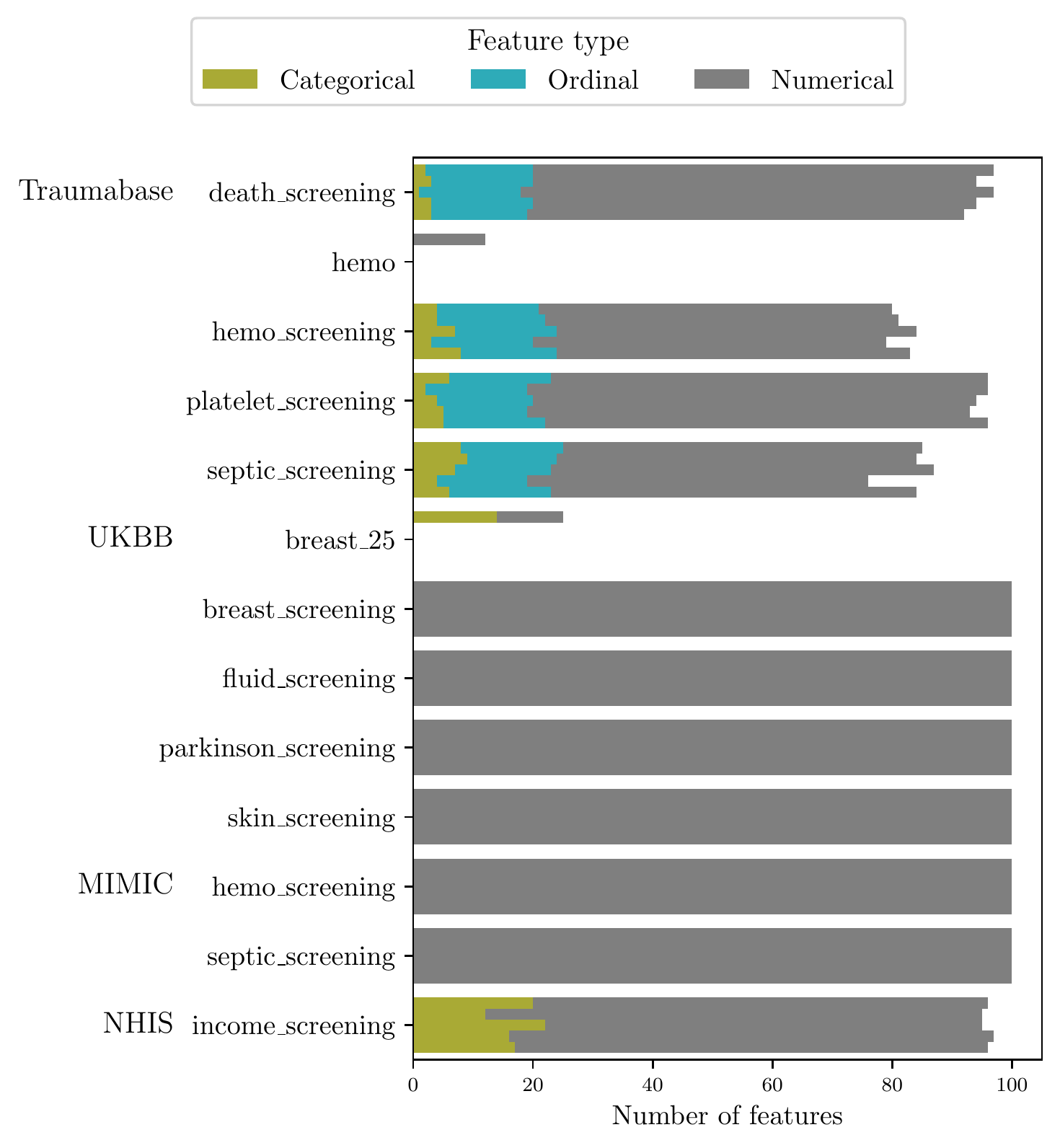}
    \label{fig:data:types}
\end{figure}

Regardless of missing-values handling, gradient-boosted trees
predict significantly better than linear models (Supplementary Table~\ref{tab:wilcoxon:gbt-vs-linear}). Tree-based models
excel on categorical or ordinal features, however these are only a
minority of the features of the databases studied (Supplementary Figure~\ref{fig:data:types}). Hence the good performance of gradient-boosted
trees probably reveals non-linear mechanisms in the data. Note that the
smallest database that we explored has a sample size of $n$=2\,500. For
much smaller data, the simplest model --the linear model-- may be the
best choice.

\subsubsection{The missingness is informative}

For imputation-based pipelines, prediction significantly improves
with the missingness mask added as input features. This suggests that the
missingness is informative, which is often the case in health databases
\citep{agniel2018biases,madden2016missing}. Hence for all health databases studied, either
the covariates are Missing Not At Random (MNAR) or the outcome to predict
depends on the missingness. Either cases fall outside of the theoretical
framework that grounds the validity of statistical analysis using
imputation \citep{Rubin1976,josse2019consistency}.
The empirical results also confirm that the practice of adding the mask
as input allows to harness the predictive information in missing data
patterns \citep{Sperrin2020}, otherwise hidden in the imputed data and much more difficult to recover.

\subsubsection{Features with high missing rates are also important.}
\begin{figure*}
    \centering
    \includegraphics[width=0.97\linewidth]{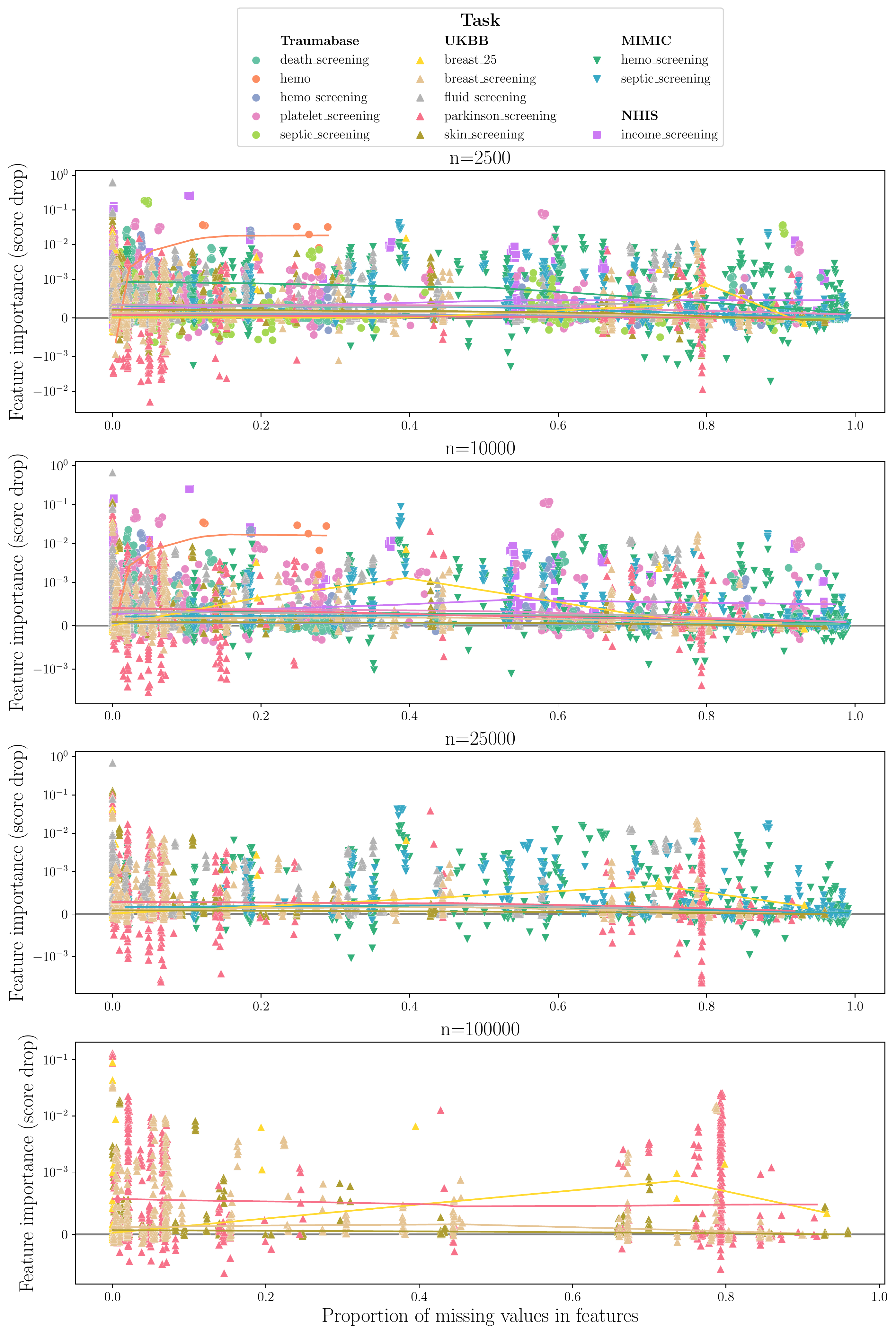}
    \caption{\textbf{Feature importance versus proportion of missing values.} Importance is measured as the drop in score when randomly permuting the considered feature. Each feature is permuted 10 times and its importance is taken as the average drop in score. Score drops are also averaged across folds. Local regressions (LOWESS) are plotted for each task to better visualize trends.}
    \label{fig:data:importance}
\end{figure*}

Within each task, the missing rate per feature varies over a wide
spectrum (see Supplementary Figure~\ref{fig:data:mv}). We checked that
features' missing rates and predictive importance were not associated.
For this, we measured permutation features: the drop in a model score
after shuffling a feature, thereby cancelling its contribution to the
model performance. We ran this experiment for each task and each feature
using scikit-learn's implementation (see Supplementary
Table~\ref{tab:methods:sklearn}). We found
no association between a feature's missing rate and its importance
(Supplementary Figure~\ref{fig:data:importance}). Predictions do not only rely on features with few missing values. Moreover, even features with a very high level of missing values (for example $>80\%$) seem to be as important as the others. This highlights the fact that it is worth making the effort of learning with incomplete features, even when they have a high missing rate.

\subsubsection{Imputation may benefit robustness or interpretability}

A good imputation may bring the benefit of recovering a meaningful
missing value, reflecting a biological or clinical reality rather than
operational constraints. For instance, the weight of a patient may be measured
upon scheduled admission to a hospital but not at the emergency
department. A predictive model based on an imputed underlying value may
lend itself better to mechanistic interpretation than a model implicitly
capturing missingness such as MIA.
In addition, using missingness to drive prediction may be more fragile,
\eg{} to changes in the operational process.
In such a case, shifts in the missing data patterns
should be closely monitored \citep{Sperrin2020,van2020cautionary,groenwold2020informative} as they could seriously
alter prediction performance.
Indeed, machine-learning models building their predictions on
``shortcuts'' in the data --not directly related to outcome of interest
but rather to the acquisition-- sometimes generalize less well to
new hospitals \citep{degrave2021ai}. Nevertheless, in health the mere
presence of a measure, such as a colonoscopy, is often an indication in
itself.

\subsection{Limitations and further work}

\subsubsection{Limitations: not all differences are significant}

Relative performance of approaches varies across datasets, which is
not surprising as no prediction model is expected to dominate on all
data. The diversity of the datasets and the statistical analysis
grounds the generality of the findings. Yet, not all differences are
significant at large sample sizes. This lack of significance can simply
be explained because of a small statistical power of the benchmark as only a few datasets are available to test these very large sample sizes settings (only 4 tasks at the $n$=100\,000 size).

More datasets would probably have made more differences significant. Yet, the benchmarks presented here already incurred large computational costs, due to the nested cross-validation: about 520\,000 CPU hours. Also, the findings build upon 13 different tasks, markedly more than the typical machine-learning benchmark: only 6\% of empirical results published at NeurIPS and 8\% ICLR (both leading machine-learning venues) build upon more than 10 datasets \citep{Bouthillier2020}.

\subsubsection{Limitations: imputation quality is not assessed}
All the conclusions of this study pertain to prediction and do not allow
us to conclude on imputation's ability to accurately reconstruct missing
values. The focus of our study is indeed on prediction.

\subsubsection{Further work: more benchmarks would be interesting, and
costly}

To limit computation costs and mimic typical usage, no hyper-parameters
tuning was performed on the parameters of the imputers. Recently,
software tools have been introduced to perform model selection on
imputation jointly with the supervised step
\citep{jarrett2021clairvoyance,nadia_R}. Further evaluation could
quantify how much gains are brought by such joint model selection, though
it would need sizable computational resources.

Further work could test more supervised learning models. The motivation
of the present study was not to find the absolute best pipeline, but
rather to understand compromises that hold across datasets and are
readily usable.

\subsection{Conclusion}

Extensive benchmarking on health databases reveals trends in the performance
of methods to build predictive models handling missing values.
First, directly incorporating missing values in tree-based models with MIA,
gives a small but systematic improvement in prediction performance over
prior imputation. Second, the computational cost of imputation using MICE or
KNN becomes intractable for large datasets. Third, gradient-boosted
trees give better predictions than linear models.
Fourth, Bagging increases predictive performance
but with a severe computational cost.
Fifth, good imputation does not imply good prediction as both have
different tradeoffs. Finally,
the experiments reveal that the missingness is informative.
Overall, a novel message of this benchmark is that for building predictive
models, supervised learning directly handling missing values should be
considered, beyond imputation.



\section{Potential implications}

This work suggests a departure from current practices: supervised learning
directly handling missing values can be preferable to imputation. In
particular, classic conditional expectation methods can be
computationally intractable both in terms of time and memory on large
datasets. Constant imputation with the mask also performs well with
little costs.

\section{Detailed benchmarking methodology}
\label{sec:detailed-methods}

\subsection{Experiment}
We selected four real databases with missing values described in \textit{\nameref{sec:datadescription}}. From them we defined empirically 13 prediction tasks – that is a set of input features and an outcome to predict – with the intent of covering as diverse use cases as possible: regressions, classifications, diverse outcomes, diverse feature types (numerical, ordinal and categorical). We sub-sampled the datasets to study 4 sizes: 2\,500, 10\,000, 25\,000 and 100\,000 samples. We selected a subset of features from the databases for each prediction task using two approaches. Manually selecting or defining features based on articles or automatically selecting 100 encoded features using an univariate ANOVA selection. We often used the later because it has the advantage of not requiring expert knowledge to define the features. Manual selection keeps fewer features than our automated selection. Note that we one-hot encoded categorical features before selecting 100 encoded features with ANOVA. Less than 100 non-encoded features may thus be involved in the task. The ANOVA is fitted on one third of the samples and the two remaining thirds are kept for fitting and evaluating the methods. To reduce bias induced by the choice of subset on which is fit the ANOVA, we ran 5 trials in which the subset is each time redrawn and average the scores and times. Task having their features manually selected are given the whole samples and only 1 trial is performed. Each of the 12 methods is given the exact same features and cross-validation folds.\\
The next step consists in benchmarking the 12 methods of Table~\ref{tab:methods:tree} on the defined prediction tasks. We used the implementation from scikit-learn \citep{pedregosa2011scikit} for all methods (see Supplementary Table~\ref{tab:methods:sklearn}).
Two nested cross-validations are used.
The outer one yields 5 training and test sets. On each training set, we perform a cross-validated hyper-parameter search –the inner cross-validation– and select the best hyper-parameters. We evaluate the best model on the respective test set. We assess the quality of the prediction with a coefficient of determination for regressions and the area under the ROC curve for classification.
We average the scores obtained on the 5 test sets of the outer cross-validation to give the final score.
Finally, we compare averaged prediction scores one to each other.\\
We also monitored training and imputation times to add time concerns to our analysis. A very detailed description of the experimental method is available on protocols.io \citep{Perez-Lebel2022-protocol}. A link to the code of the experiments is given in \textit{\nameref{sec:availability:code}}.

\subsection{Plotting method}
\label{sec:plotting-method}
\subsubsection{Figures on prediction scores}
The experiment gives one prediction score per fold, per trial, per task, per method, per size. Cross-validations aggregate and average scores across the folds and trials resulting in an average score for each of the (task, method, size). For each one of the pairs (task, size), we computed a reference score by averaging the scores obtained by the 12 methods on the corresponding task and size.
The plotted metric is what we called the relative prediction score - that is the deviation of the prediction score from the reference score - for each of the (task, method, size).
We created one box plot for each of the 4 sizes with the same structure: the relative prediction score on the x-axis and the 12 methods on the y-axis. Each is overlaid with a scatter plot plotting the relative prediction score per (task, method, size). The scatter plot shares its x-axis with the box plot.
On the y-axis however, each dot is given a y coordinate according to its method and database so that scores coming from a same method and database are plotted on the same horizontal line.

\subsubsection{Figures on computation time}
Computational time plots follow the same structure. The metric of interest is now the total training time. It includes imputation time and the full hyper-parameters tuning time. It is evaluated using computer's process time instead of wall-clock time.
The total training time of MIA is taken as reference time for each (task, size). The relative total training time is computed by dividing by the reference time. The x-scale is logarithmic to better apprehend comparison on large scales.

\section{Availability of source code and requirements}
\label{sec:availability:code}
\begin{itemize}
\item Project name: Benchmarking missing-values approaches for predictive models on health databases.
\item Project home page: \url{https://github.com/aperezlebel/benchmark_mv_approaches}
\item Operating system: Platform independent
\item Programming language: Python 3.7.6
\item Other requirements: all requirements are listed in the \texttt{requirements.txt} file of the repository.
\item License: MIT
\end{itemize}

\section{Data Availability}
\label{sec:availability:data}
All supporting data and materials are available in the GigaScience GigaDB database \citep{Perez-Lebel2022}. The datasets supporting the results of this article are available at the following URL.
\begin{itemize}
    \item \textbf{Traumabase}, by contacting the team at \url{http://www.traumabase.eu/en_US/contact}.
    \item \textbf{UKBB}: upon application at \url{https://www.ukbiobank.ac.uk/register-apply/}.
    \item \textbf{MIMIC-III}: upon application at \url{https://mimic.physionet.org/gettingstarted/access/}.
    \item \textbf{NHIS} freely available at \url{https://www.cdc.gov/nchs/nhis/nhis_2017_data_release.htm}.
\end{itemize}

A thorough description of the protocols of the experiments conducted in this article is available on protocols.io \citep{Perez-Lebel2022-protocol}.






\section{Declarations}
\subsection{List of abbreviations}
The following abbreviations were used in this article:
\begin{itemize}
    \item ANOVA: Analysis Of Variance.
    \item AUC: Area Under the Curve.
    \item KNN: K-Nearest Neighbors.
    \item MAR: Missing At Random
    \item MIA: Missing Incorporated in Attribute.
    \item MNAR: Missing Not At Random
    \item MIMIC: Medical Information Mart for Intensive Care.
    \item NHIS: National Health Interview Survey.
    \item LOWESS : Locally Weighted Scatterplot Smoothing.
    \item UKBB: UK Biobank.
\end{itemize}

\subsection{Ethical Approval}
Only secondary data use was involved in this work. Each one of the
databases has previously been approved by the Ethical Review Board.
Access to each database was obtained according to the corresponding rules
and authorization, when applicable (MIMIC, Traumabase, UKBB).

\subsection{Consent for publication}
Not applicable.

\subsection{Competing Interests}
The authors declare that they have no competing interests.

\subsection{Funding}

Gaël Varoquaux, Julie Josse, and Marine Le Morvan acknowledge funding from DataAI under the MissingBigData grant. Gaël Varoquaux and Marine Le Morvan acknowledge funding from ANR under the DirtyData grant (ANR-17-CE23-0018). Julie Josse acknowledges funding from ANR under the MUSE grant (ANR-16-IDEX-0006). Alexandre Perez-Lebel was partially funded by a Mitacs Globalink Research Award for research in Canada. Jean-Baptiste Poline is partially funded by the National Institutes of Health (NIH) NIH-NIBIB P41 EB019936 (ReproNim)  NIH-NIMH R01 MH083320 (CANDIShare) and NIH RF1 MH120021 (NIDM), the National Institute Of Mental Health of the NIH under Award Number R01MH096906 (Neurosynth), as well as the Canada First Research Excellence Fund, awarded to McGill University for the Healthy Brains for Healthy Lives initiative and the Brain Canada Foundation with support from Health Canada.

\subsection{Author's Contributions}
Following the CRediT Contributor Roles Taxonomy:
\begin{itemize}
    \item Alexandre Perez-Lebel: Investigation, Data curation, Formal Analysis, Writing - original draft, Writing - review \& editing.
    \item Gaël Varoquaux: Funding acquisition, Conceptualization, Supervision, Writing - review \& editing, Validation.
    \item Marine Le Morvan: Supervision, Writing - review \& editing, Validation.
    \item Julie Josse: Writing - review \& editing.
    \item Jean-Baptiste Poline: Funding acquisition, Supervision, Writing - review \& editing, Validation.
\end{itemize}

\section{Acknowledgements}
Not applicable.





\section{Authors' information}
Alexandre Perez-Lebel is a PhD student in machine learning at Inria.
This work started as an internship at the Montreal Neurological Institute of McGill University and extended into the early months of the PhD program.

\bibliography{refs}

\clearpage

\begin{figure*}[t]
    \begin{subfigure}{\linewidth}
    \subcaption{{\normalsize\textbf{Prediction performance}}}
    \label{fig:finding:linear:score}
    \includegraphics[width=\textwidth]{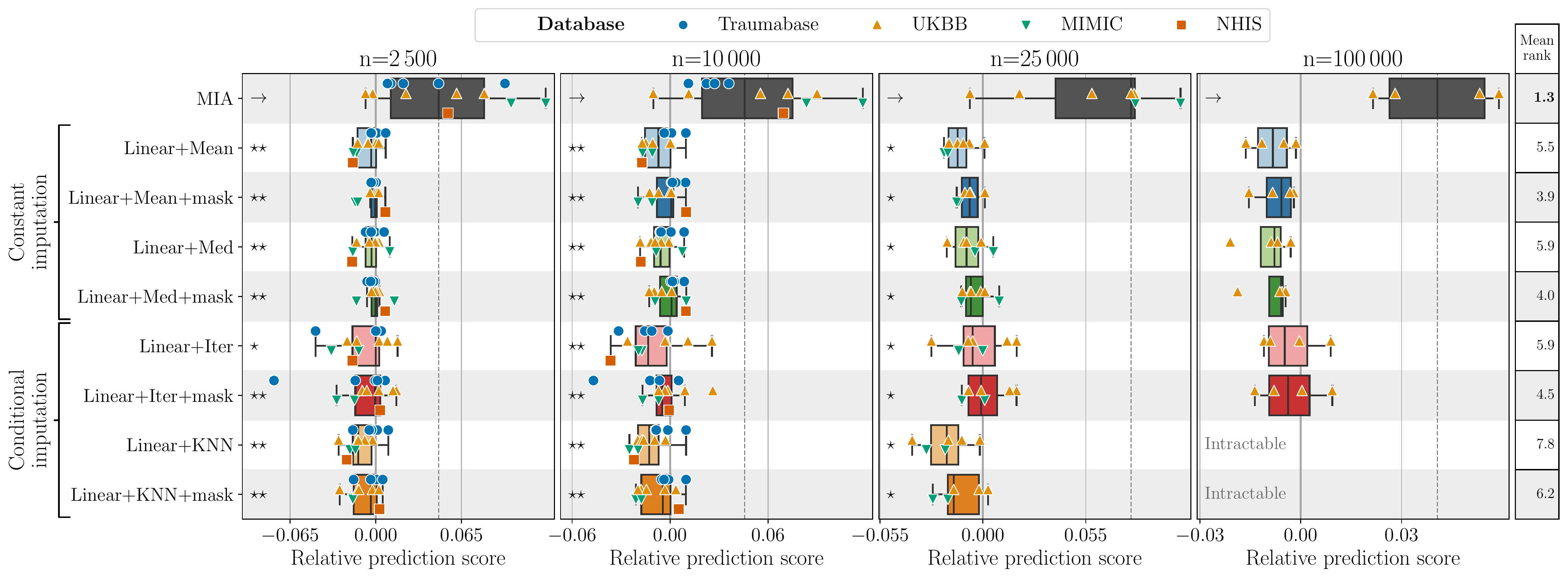}
    \end{subfigure}

    \medskip
    \begin{subfigure}{\linewidth}
    \subcaption{{\normalsize\textbf{Computational time}}}
    \label{fig:finding:linear:time}
    \includegraphics[width=\textwidth]{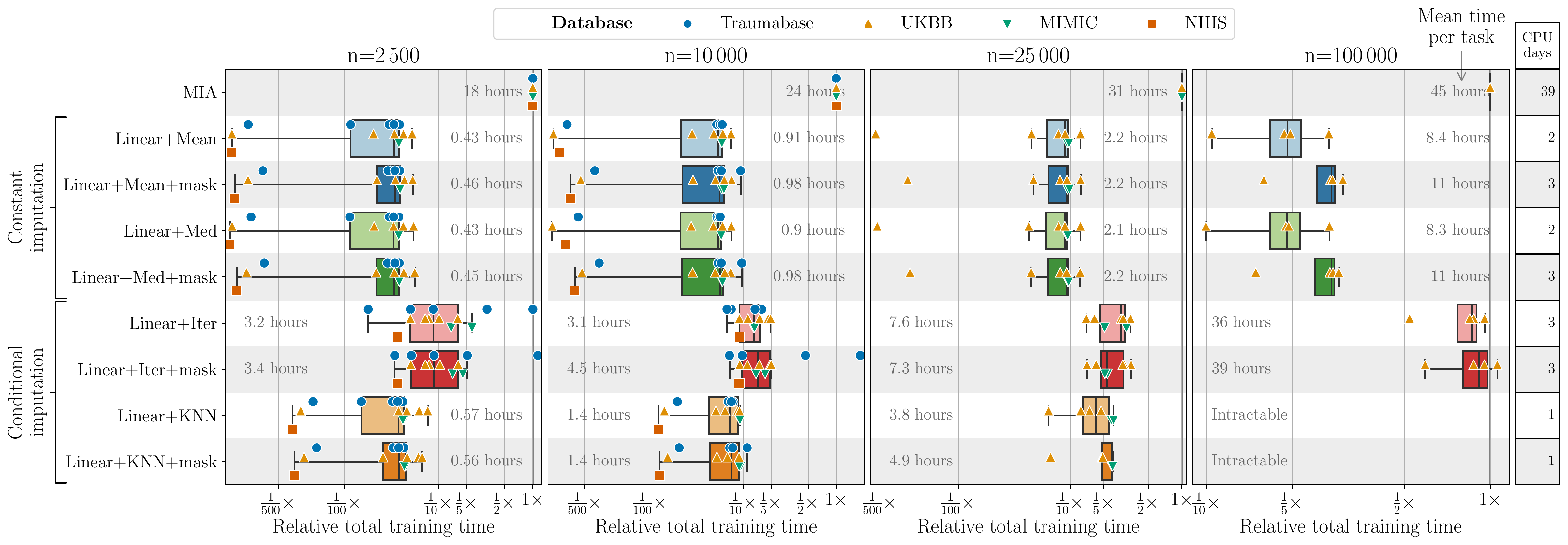}
    \end{subfigure}
    \caption{\textbf{Supplementary results: linear and gradient-boosted trees models.}
    Comparison of prediction performance and training times across the 9 methods (linear models and gradient boosting trees, see Supplementary Table~\ref{tab:methods:linear}) for 13 prediction tasks spread over 4 databases, and for 4 sizes of dataset (2\,500, 10\,000, 25\,000 and 100\,000 samples). For linear models, ridge is used for regressions and logistic regression for classifications.
    For each of the tasks and sizes, we computed a reference score by averaging the scores obtained by the 9 methods on the corresponding task and size. The relative prediction score of a method on a task and size is the deviation of the prediction score from the reference score of this task and size. For computational time, the total training time comprises imputation and tuning times and is given relative to the one of MIA for each task and size. More details on how these plots were created are given in the \textit{\nameref{sec:detailed-methods}} section.
    The significance is assessed with a one-sided Wilcoxon signed-rank test with MIA taken as reference (see Supplementary Table~\ref{tab:wilcoxon:gbt-vs-linear}). Methods which performed significantly poorer (resp. better) at the 0.05 level are marked with "$\star$" (resp. "$\star(>)$") and "$\star\star$" (resp. "$\star\star(>)$") for Bonferroni-corrected levels.
    Two tables give the overall average ranks and the total number of CPU days for each method, all tasks and sizes combined. The average number of CPU hours \textit{per task} required to evaluate each method is given on each line. Detailed scores and ranks broken out by tasks are given in Supplementary Table~\ref{tab:linear:scores-ranks}. Notice that KNN and KNN+mask were intractable at $n$=100\,000 due to their memory footprint of $\mathcal{O}(n^2)$.
    }
    \label{fig:finding:linear}
\end{figure*}


\pagebreak
\section{Appendix}


\subsection{Supplementary experiment: linear models or trees?}

\label{sec:linear_models}

\subsubsection{Protocol}
This supplementary experiment uses the same pipeline as the main experiment except that imputation is paired with linear models instead of boosted trees as summarized in Supplementary Table~\ref{tab:methods:linear}. We used ridge for regressions, and $\ell_2$-penalized logistic regression for classifications.


\subsubsection{Findings: trees with MIA improve upon linear models}
MIA with boosted trees outperforms all 8 combinations of imputers with linear models, on every size and every database. Supplementary Figure~\ref{fig:finding:linear:score} shows that MIA obtained the best average rank of 1.3 far ahead of other methods. The following ones are Linear+Mean+mask, Linear+Med+mask and Linear+Iter+mask with a rank of 3.9, 4.0 and 4.5 respectively. The one-sided Wilcoxon signed-rank test confirms this claim. Supplementary Table~\ref{tab:wilcoxon} shows that MIA with boosted trees is significantly better than every linear methods on the first two sizes even at the Bonferroni-corrected level. The null hypothesis of the Friedman test is rejected below the 0.05 level except for the last size as shown on Supplementary Table~\ref{tab:linear:friedman}. Thus methods are not equivalent for the first three sizes. The Nemenyi test on Supplementary Figure~\ref{fig:linear:nemenyi} confirms that results are not significant for the larger size.\\
Moreover, we were expecting the mean and median imputations to give bad results being paired with linear models as shown in \citet{Morvan2020a}. Not only these results confirm our expectations, but they also show that non-constant imputation models give similar results when paired with a linear model. As before, masked versions perform slightly better than their no-mask counterpart.\\
However, gradient-boosted trees with MIA are a lot slower than imputation with linear models. Supplementary Figure~\ref{fig:finding:linear:time} shows that boosted trees with MIA is up to 500 times slower than constant imputations with linear models. Also, conditional imputation leads to slower computations than mean and median imputation. Given the low gain obtained against mean and median imputation, they are of limited interest.\\
The main takeaway is the outperformance in score of MIA with gradient-boosted trees over imputation with linear models when it comes to handling missing values. This outperformance comes with a cost: a much longer computation time.

\begin{table}[t!]
    \caption{\textbf{Methods compared in the supplementary experiment.}}
    \label{tab:methods:linear}
    \setlength\tabcolsep{8pt}
    \begin{tabularx}{\linewidth}{l l l l l}
    \toprule
    \makecell{In-article name} & \makecell{Imputer} & \makecell{Mask} & \makecell{Predictive model} \\
    \midrule
    Boosted trees+MIA & - & - & Boosted trees \\
    Linear+Mean & Mean & No & Ridge/Logit \\
    Linear+Mean+mask & Mean & Yes & Ridge/Logit \\
    Linear+Med & Median & No & Ridge/Logit \\
    Linear+Med+mask & Median & Yes & Ridge/Logit \\
    Linear+Iter & Iterative & No & Ridge/Logit \\
    Linear+Iter+mask & Iterative & Yes & Ridge/Logit \\
    Linear+KNN & KNN & No & Ridge/Logit \\
    Linear+KNN+mask & KNN & Yes & Ridge/Logit \\
    \bottomrule
\end{tabularx}

\end{table}

\subsection{Significance tests}

In the following paragraphs, we took the notations and formulations of \citet{Demsar2006}. We consider $k$ algorithms and $N$ datasets. We note $r_i^j$ the rank of the $j$-th algorithm on the $i$-th dataset. Note $R_j = \frac{1}{N}\sum_i r_i^j$ the average rank.

\paragraph{Friedman test}
 The Friedman statistic $\chi_F^2$ is distributed according to a chi-square distribution with $k-1$ degrees of freedom.
\begin{equation}
    \chi_F^2 = \frac{12N}{k(k+1)}\bra{\sum_jR_j^2 - \frac{k(k+1)^2}{4}}
    \label{eq:test:chi2}
\end{equation}
\citet{Iman1980} derived a less conservative statistic $F_F$ which is distributed according to the F-distribution with $k-1$ and $(k-1)(N-1)$ degrees of freedom.
\begin{equation}
    F_F = \frac{(N-1)\chi^2_F}{N(k-1)-\chi^2_F}
    \label{eq:test:ff}
\end{equation}

\noindent Both statistics (\ref{eq:test:chi2}) and (\ref{eq:test:ff}) are given on Supplementary Table~\ref{tab:friedman} with their associated p-values for the 2 sets of methods and the 4 sizes of dataset.

\begin{table}[H]
    \caption{\textbf{Friedman test}, correction by Iman and Davenport and Nemenyi test. CD is the critical distance and N the number of tasks for each size.}
    \label{tab:friedman}

    \centering
    \setlength{\tabcolsep}{5pt}
    \begin{subtable}{0.48\textwidth}
        \subcaption{Tree-based methods of Table~\ref{tab:methods:tree}.}
        \begin{tabularx}{\textwidth}{lllllll}
\toprule
{} & \hphantom{-}$\chi^2_F$ & \hphantom{-}$\chi^2_F$ p-value & \hphantom{-}$F_F$ & \hphantom{-}$F_F$ p-value &   \hphantom{-}CD &   \hphantom{-}N \\
Size   &                        &                                &                   &                           &                  &                 \\
\midrule
2500   &         \hphantom{-}73 &            \hphantom{-}9.6e-11 &    \hphantom{-}12 &       \hphantom{-}7.4e-16 &  \hphantom{-}4.6 &  \hphantom{-}13 \\
10000  &         \hphantom{-}76 &            \hphantom{-}2.3e-11 &    \hphantom{-}15 &       \hphantom{-}6.1e-18 &  \hphantom{-}4.8 &  \hphantom{-}12 \\
25000  &         \hphantom{-}30 &            \hphantom{-}2.5e-03 &   \hphantom{-}3.9 &       \hphantom{-}2.4e-04 &  \hphantom{-}6.3 &   \hphantom{-}7 \\
100000 &         \hphantom{-}10 &               \hphantom{-}0.43 &   \hphantom{-}1.2 &          \hphantom{-}0.35 &  \hphantom{-}6.8 &   \hphantom{-}4 \\
\bottomrule
\end{tabularx}

        \label{tab:trees:friedman}
    \end{subtable}%
\end{table}
\bigskip
\begin{table}
    \ContinuedFloat
    \begin{subtable}{0.48\textwidth}
        \setlength{\tabcolsep}{5pt}
        \subcaption{Boosted-trees and linear methods of Supplementary Table~\ref{tab:methods:linear}.}
        \begin{tabularx}{\textwidth}{lllllll}
\toprule
{} & \hphantom{-}$\chi^2_F$ & \hphantom{-}$\chi^2_F$ p-value & \hphantom{-}$F_F$ & \hphantom{-}$F_F$ p-value &   \hphantom{-}CD &   \hphantom{-}N \\
Size   &                        &                                &                   &                           &                  &                 \\
\midrule
2500   &         \hphantom{-}41 &            \hphantom{-}5.1e-06 &   \hphantom{-}7.8 &       \hphantom{-}5.2e-08 &  \hphantom{-}3.3 &  \hphantom{-}13 \\
10000  &         \hphantom{-}50 &            \hphantom{-}9.7e-08 &    \hphantom{-}12 &       \hphantom{-}1.7e-11 &  \hphantom{-}3.5 &  \hphantom{-}12 \\
25000  &         \hphantom{-}23 &            \hphantom{-}5.6e-03 &   \hphantom{-}4.3 &       \hphantom{-}6.2e-04 &  \hphantom{-}4.5 &   \hphantom{-}7 \\
100000 &                    -19 &                  \hphantom{-}1 &              -1.1 &             \hphantom{-}1 &    \hphantom{-}6 &   \hphantom{-}4 \\
\bottomrule
\end{tabularx}

        \label{tab:linear:friedman}
    \end{subtable}
\end{table}

\paragraph{Nemenyi test}
Once the Friedman test is rejected, the Nemenyi test can be applied. It provides a critical difference CD which is the minimal difference between the average ranks of two algorithms for them to be significantly different.
\begin{equation}
    \label{eq:test:cd}
    CD = q_{\alpha}\sqrt{\frac{k(k+1)}{6N}}
\end{equation}
Values of $q_{\alpha}$ are given in Table~5 of \citet{Demsar2006}. Values of critical differences for the 2 sets of methods and the 4 sizes of dataset are given in Supplementary Table~\ref{tab:friedman}.

\paragraph{Wilcoxon signed-rank test}
To compute the one-sided Wilcoxon signed-rank test, we used the \texttt{wilcoxon} function of the \texttt{scipy.stats} module between the 13 average scores of MIA against the ones of every other methods. Resulting p-values are given in Supplementary Table~\ref{tab:wilcoxon} for the 4 sizes of dataset.

\begin{table}[H]
    \caption{\textbf{One-sided Wilcoxon signed-rank test.} p-values of the one-sided right test on the difference of score between MIA and every other method for Table~\ref{tab:wilcoxon:mia-vs-rest}, and between gradient-boosted trees and linear models for Table~\ref{tab:wilcoxon:gbt-vs-linear}. p-values below the 0.05 level are marked with ${}^{\star}$. p-values below the Bonferroni corrected level are marked with ${}^{\star\star}$. When the reversed test (\ie{} one-sided left) is significant instead, p-values are marked with ${}^{\star(>)}$ and ${}^{\star\star(>)}$ following the same rule.}
    \label{tab:wilcoxon}

    \centering
    \setlength{\tabcolsep}{2.9pt}
    \begin{subtable}{0.483\textwidth}
        \subcaption{\textbf{MIA vs imputation.} Bonferroni level: $0.05/19 = 2.6\times10^{-3}$. Rejecting the null hypothesis means MIA performed better than the compared method.}
        \begin{tabularx}{\textwidth}{lllll}
\toprule
Size &                         2500   &                         10000  &                    25000  &   100000 \\
Method               &                                &                                &                           &          \\
\midrule
Mean                 &  $\text{1.2e-03}^{\star\star}$ &       $\text{4.6e-02}^{\star}$ &  $\text{2.3e-02}^{\star}$ &  6.2e-02 \\
Mean+mask            &       $\text{4.0e-02}^{\star}$ &                        2.3e-01 &                   1.5e-01 &  6.2e-02 \\
Median               &       $\text{5.2e-03}^{\star}$ &  $\text{1.7e-03}^{\star\star}$ &  $\text{2.3e-02}^{\star}$ &  6.2e-02 \\
Median+mask          &       $\text{4.0e-02}^{\star}$ &                        2.1e-01 &                   1.5e-01 &  1.2e-01 \\
Iterative            &       $\text{5.2e-03}^{\star}$ &       $\text{3.2e-02}^{\star}$ &  $\text{3.9e-02}^{\star}$ &  6.2e-02 \\
Iterative+mask       &       $\text{2.4e-02}^{\star}$ &                        2.1e-01 &                   4.7e-01 &  6.2e-02 \\
KNN                  &  $\text{1.2e-04}^{\star\star}$ &  $\text{2.4e-04}^{\star\star}$ &  $\text{3.1e-02}^{\star}$ &          \\
KNN+mask             &  $\text{1.2e-04}^{\star\star}$ &  $\text{7.3e-04}^{\star\star}$ &  $\text{3.1e-02}^{\star}$ &          \\
\midrule MI          &                        8.1e-01 &                        6.6e-01 &                   3.4e-01 &  1.2e-01 \\
MI+mask              &    $\text{9.9e-01}^{\star(>)}$ &    $\text{9.6e-01}^{\star(>)}$ &                   9.2e-01 &  4.4e-01 \\
MIA+bagging          &    $\text{9.7e-01}^{\star(>)}$ &                        9.4e-01 &                   7.7e-01 &  3.1e-01 \\
\midrule Linear+Mean &  $\text{6.1e-04}^{\star\star}$ &  $\text{4.9e-04}^{\star\star}$ &  $\text{7.8e-03}^{\star}$ &  6.2e-02 \\
Linear+Mean+mask     &  $\text{8.5e-04}^{\star\star}$ &  $\text{7.3e-04}^{\star\star}$ &  $\text{1.6e-02}^{\star}$ &  6.2e-02 \\
Linear+Med           &  $\text{6.1e-04}^{\star\star}$ &  $\text{4.9e-04}^{\star\star}$ &  $\text{7.8e-03}^{\star}$ &  6.2e-02 \\
Linear+Med+mask      &  $\text{6.1e-04}^{\star\star}$ &  $\text{4.9e-04}^{\star\star}$ &  $\text{1.6e-02}^{\star}$ &  6.2e-02 \\
Linear+Iter          &       $\text{3.1e-03}^{\star}$ &  $\text{1.2e-03}^{\star\star}$ &  $\text{1.6e-02}^{\star}$ &  6.2e-02 \\
Linear+Iter+mask     &  $\text{2.3e-03}^{\star\star}$ &  $\text{1.2e-03}^{\star\star}$ &  $\text{1.6e-02}^{\star}$ &  6.2e-02 \\
Linear+KNN           &  $\text{1.2e-04}^{\star\star}$ &  $\text{2.4e-04}^{\star\star}$ &  $\text{1.6e-02}^{\star}$ &  5.0e-01 \\
Linear+KNN+mask      &  $\text{1.2e-04}^{\star\star}$ &  $\text{2.4e-04}^{\star\star}$ &  $\text{3.1e-02}^{\star}$ &  5.0e-01 \\
\bottomrule
\end{tabularx}

        \label{tab:wilcoxon:mia-vs-rest}
    \end{subtable}%

\end{table}
\bigskip
\begin{table}
    \ContinuedFloat
    \centering
    \setlength{\tabcolsep}{5pt}
    \begin{subtable}{0.48\textwidth}
        \subcaption{\textbf{Gradient-boosted trees vs linear models.} Bonferroni level: $0.05/8 = 6.25\times10^{-3}$. Rejecting the null hypothesis means gradient-boosted trees performed better than linear models for the given imputer.}
        \begin{tabularx}{\textwidth}{lllll}
\toprule
Size &                         2500   &                         10000  &                    25000  &   100000 \\
Imputer        &                                &                                &                           &          \\
\midrule
Mean           &  $\text{1.2e-03}^{\star\star}$ &  $\text{4.9e-04}^{\star\star}$ &  $\text{1.6e-02}^{\star}$ &  6.2e-02 \\
Mean+mask      &  $\text{1.7e-03}^{\star\star}$ &  $\text{4.9e-04}^{\star\star}$ &  $\text{1.6e-02}^{\star}$ &  6.2e-02 \\
Median         &  $\text{6.1e-04}^{\star\star}$ &  $\text{7.3e-04}^{\star\star}$ &  $\text{1.6e-02}^{\star}$ &  6.2e-02 \\
Median+mask    &  $\text{8.5e-04}^{\star\star}$ &  $\text{2.4e-04}^{\star\star}$ &  $\text{1.6e-02}^{\star}$ &  6.2e-02 \\
Iterative      &  $\text{4.0e-03}^{\star\star}$ &  $\text{1.2e-03}^{\star\star}$ &  $\text{2.3e-02}^{\star}$ &  6.2e-02 \\
Iterative+mask &  $\text{2.3e-03}^{\star\star}$ &  $\text{1.2e-03}^{\star\star}$ &  $\text{1.6e-02}^{\star}$ &  6.2e-02 \\
KNN            &  $\text{8.5e-04}^{\star\star}$ &  $\text{7.3e-04}^{\star\star}$ &  $\text{3.1e-02}^{\star}$ &          \\
KNN+mask       &  $\text{8.5e-04}^{\star\star}$ &  $\text{4.9e-04}^{\star\star}$ &  $\text{3.1e-02}^{\star}$ &          \\
\bottomrule
\end{tabularx}

        \label{tab:wilcoxon:gbt-vs-linear}
    \end{subtable}%

\end{table}


\begin{table}[h!]
    \caption{\textbf{Scikit-learn's implementations of the methods.}}
    \label{tab:methods:sklearn}
    \setlength\tabcolsep{5.5pt}
    \begin{tabularx}{\linewidth}{l l}
    \toprule
    \makecell[l]{In-article name} & \makecell[l]{Scikit-learn's method}\\
    \midrule
    Boosted trees & \makecell[l]{\texttt{HistGradientBoostingRegressor}, \\\texttt{HistGradientBoostingClassifier}}\\
    Linear model & \makecell[l]{\texttt{Ridge}, \texttt{LogisticRegression}}\\
    Mean, Mean+mask & \texttt{SimpleImputer}\\
    Median, Median+mask & \texttt{SimpleImputer}\\
    Iterative, Iterative+mask & \texttt{IterativeImputer}\\
    KNN, KNN+mask & \texttt{KNNImputer}\\
    ANOVA selection & \makecell[l]{\texttt{f\_regression}, \texttt{f\_classif}}\\
    Permutation importance & \texttt{permutation\_importance}\\
    Bagging & \texttt{BaggingRegressor}, \texttt{BaggingClassifier}\\
    \bottomrule
\end{tabularx}

\end{table}


\begin{table}[h!]
    \setlength{\tabcolsep}{4pt}
    \caption{\textbf{Correlation between features.} Average number of ordinal and numerical features correlated to other ordinal or numerical features with an absolute correlation coefficient larger than thresholds $\{0.1, 0.2, 0.3\}$, averaged on all ordinal and numerical features of the task and expressed in percentage of the number of ordinal and numerical features in the task. For example in the task "death\_screening", a numerical or ordinal feature has an absolute correlation value greater than 0.01 with 68\% of the ordinal and numerical features of the task in average.}
    \begin{tabularx}{\linewidth}{llllll}
\toprule
    &                   &    & \multicolumn{3}{l}{Threshold} \\
    &                   &    &       0.1 &   0.2 &   0.3 \\
Database & Task & \# features &           &       &       \\
\midrule
\rotsmash{Traumabase} & death\_screening & 92 &      68\% &  41\% &  22\% \\
    & hemo & 12 &      50\% &  23\% &  12\% \\
    & hemo\_screening & 76 &      65\% &  36\% &  20\% \\
    & platelet\_screening & 90 &      67\% &  40\% &  22\% \\
    & septic\_screening & 76 &      68\% &  37\% &  18\% \\
\rule{0pt}{0.15in} UKBB & breast\_25 & 11 &      40\% &  20\% &  19\% \\
    & breast\_screening & 100 &      26\% &  12\% &   8\% \\
    & fluid\_screening & 100 &      21\% &  10\% &   6\% \\
    & parkinson\_screening & 100 &      28\% &  16\% &  11\% \\
    & skin\_screening & 100 &      24\% &  11\% &   8\% \\
\rule{0pt}{0.15in} MIMIC & hemo\_screening & 100 &      22\% &   6\% &   3\% \\
    & septic\_screening & 100 &      21\% &   6\% &   2\% \\
\rule{0pt}{0.15in} NHIS & income\_screening & 78 &      15\% &   6\% &   4\% \\
\rule{0pt}{0.2in} Average &                   & 79 &      40\% &  20\% &  12\% \\
\bottomrule
\end{tabularx}

    \label{tab:data:cor}
\end{table}

\clearpage

\begin{figure}[H]
    \begin{subfigure}{\linewidth}
        \centering
        \includegraphics[width=\linewidth]{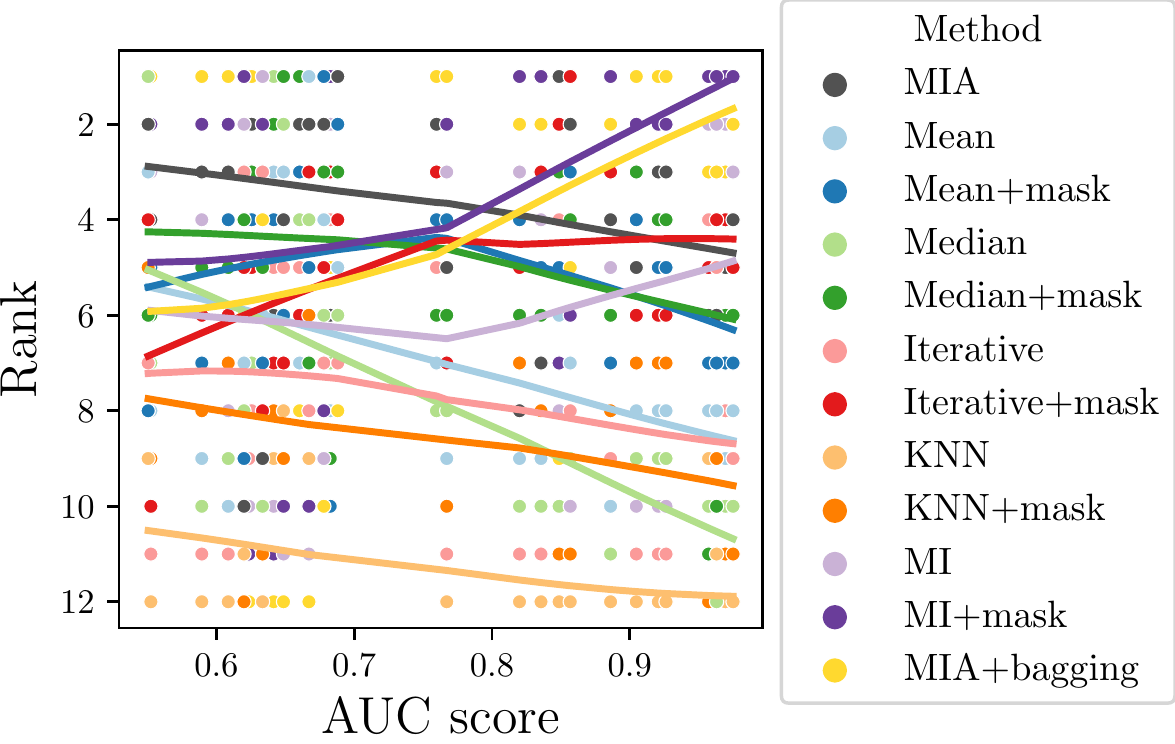}
        \caption{Classification tasks}
        \label{fig:data:difficulty:auc}
    \end{subfigure}

    \begin{subfigure}{\linewidth}
        \centering
        \includegraphics[width=\linewidth]{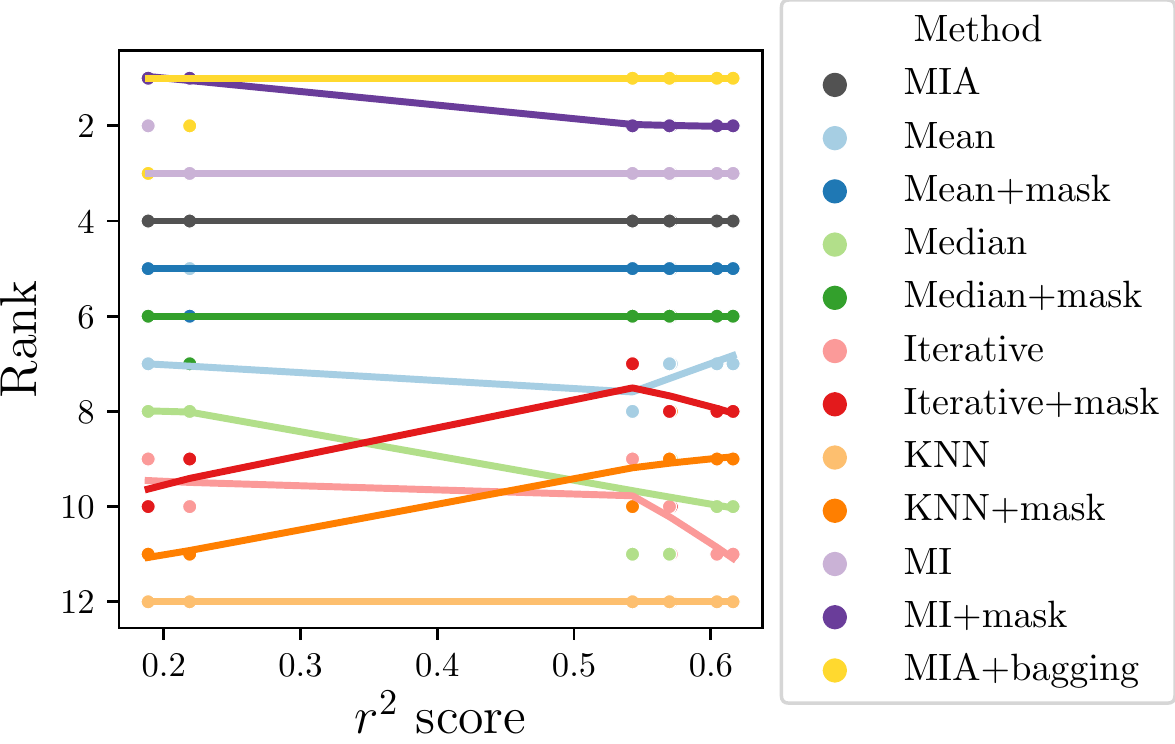}
        \caption{Regression tasks}
        \label{fig:data:difficulty:r2}
    \end{subfigure}
    \caption{\textbf{Effect of difficulty on the ranks of the methods.} For each task and size, the average score obtained by the methods is taken as a proxy of its difficulty. Local regressions (LOWESS) are plotted for each method to better visualize trends.}
    \label{fig:data:difficulty}
\end{figure}

\paragraph{Effect of tasks' difficulty on the performance of the methods}
For classification tasks, Supplementary Figure~\ref{fig:data:difficulty:auc} shows the relative performance of the methods as a function of the tasks' difficulty. Bagged methods Iterative+mask+Bagging and MIA+Bagging show a clear trend with lower (resp. higher) ranks for easier (resp. harder) methods. Also, MIA is the best performing one for harder tasks (for AUC < 0.8). Thus, the interest of MIA seems more pronounced for harder tasks. There is not enough regression tasks to observe exploitable trends on Supplementary Figure~\ref{fig:data:difficulty:r2}.

\newpage
\begin{figure*}
    \centering
    \includegraphics[width=.9\linewidth]{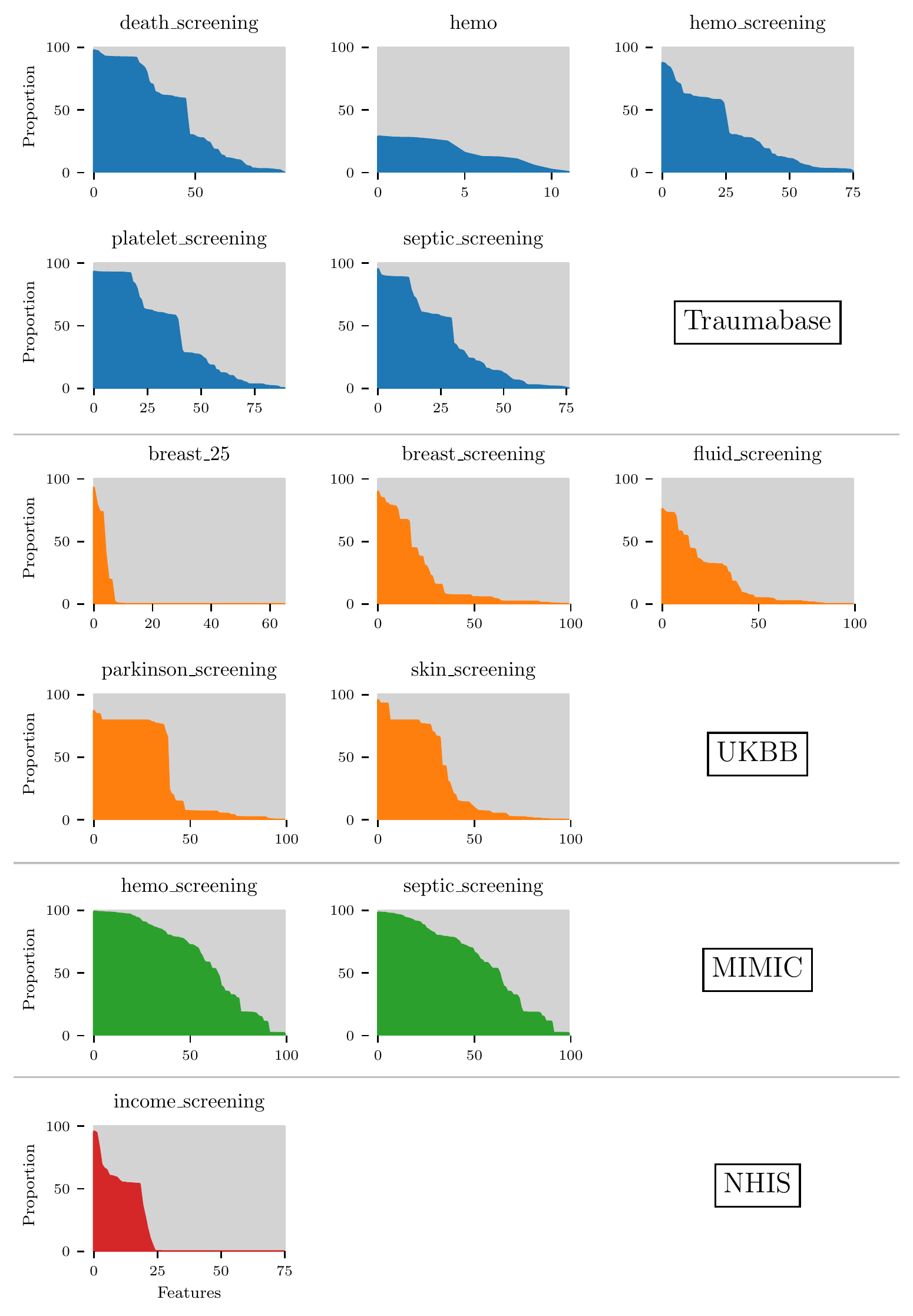}
    \caption{\textbf{Missing values distribution.} Proportion of missing values across selected encoded features for each task and for trial number 1, sorted in decreasing order. Other trials have similar proportions.}
    \label{fig:data:mv}
\end{figure*}

\newpage
\begin{figure*}
    \centering
    \includegraphics[width=0.94\linewidth]{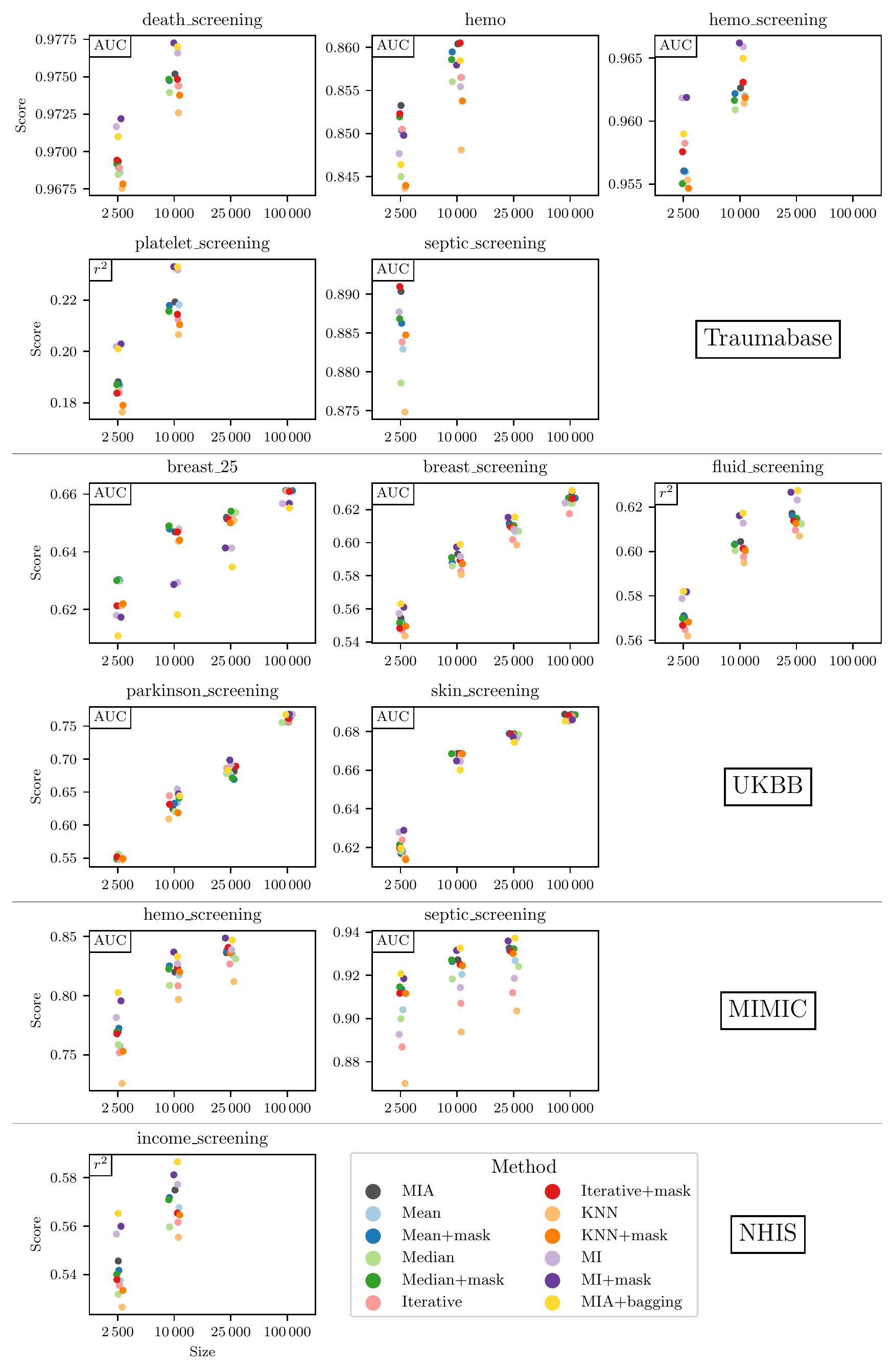}
    \caption{\textbf{Scores of the tree-based methods as a function of the training size.} Detailed scores of Figure~\ref{fig:finding:trees} broken out by task.}
    \label{fig:data:breakout}
\end{figure*}

\newpage

\begin{figure*}[t!]
    \begin{subfigure}{\linewidth}
        \centering
        \subcaption{{\normalsize\textbf{Prediction performance}}}
        \label{fig:finding:bagging:score}
        \includegraphics[width=\textwidth]{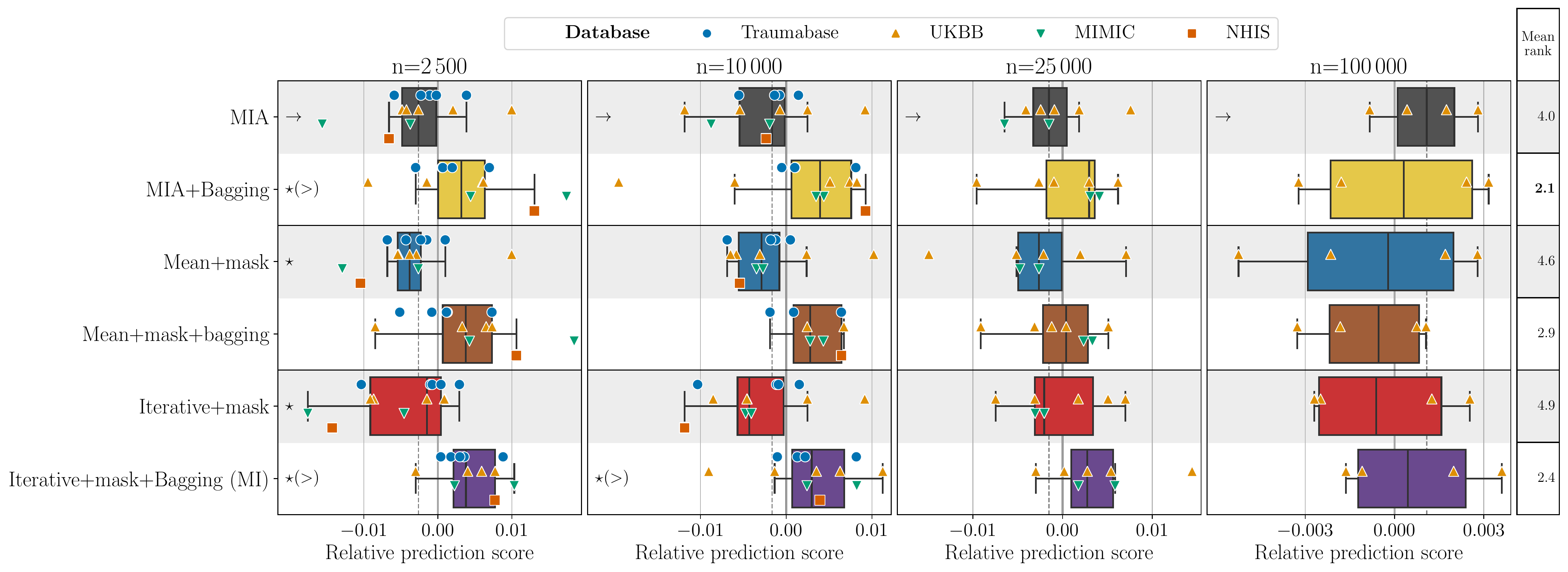}
    \end{subfigure}

    \medskip
    \begin{subfigure}{\linewidth}
        \centering
        \subcaption{{\normalsize\textbf{Computational time}}}
        \label{fig:finding:bagging:time}
        \includegraphics[width=\textwidth]{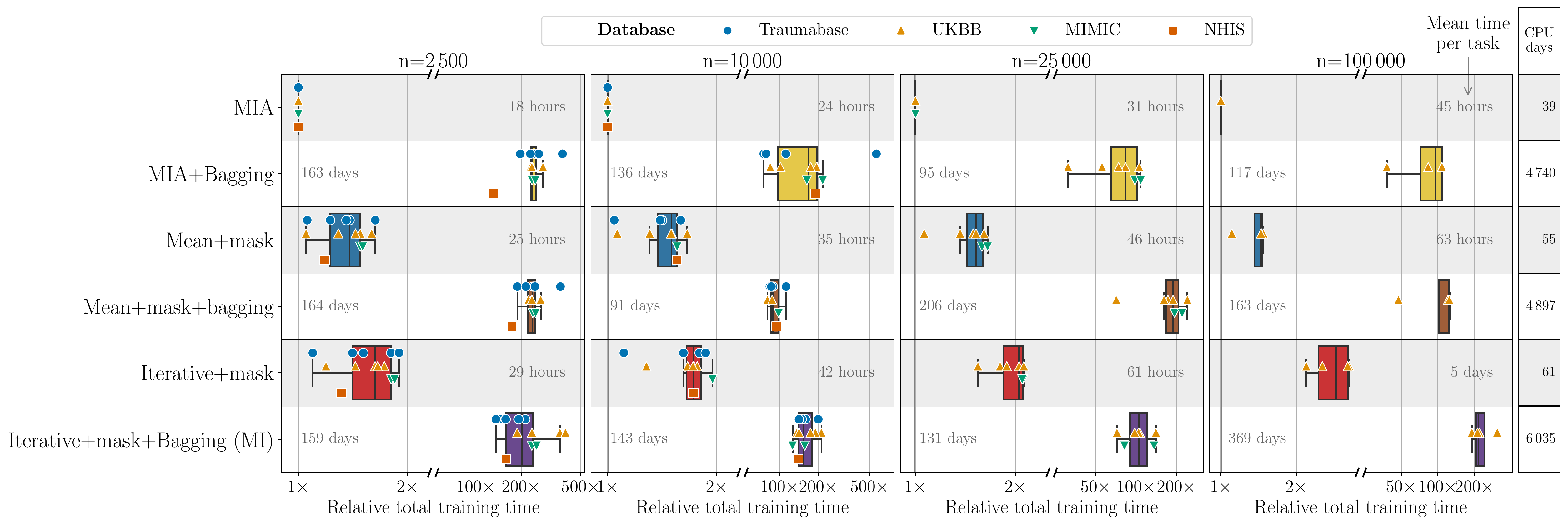}
    \end{subfigure}
    \caption{\textbf{Effect of bagging.} Comparison of prediction performance and training times between MIA, Mean+mask, Iterative+mask and their bagged version, for 13 prediction tasks spread over 4 databases, and for 4 sizes of dataset (2\,500, 10\,000, 25\,000 and 100\,000 samples). This figure is based on Figure~\ref{fig:finding:trees}, refer to caption of Figure~\ref{fig:finding:trees} for more details.}
    \label{fig:finding:bagging}
\end{figure*}

\setlength{\tabcolsep}{1.5pt}
\begin{table*}
    \tiny
    \caption{\textbf{Overview of the prediction tasks used in this article.} For selection, 'A' means ANOVA and 'M' means manual. Type 'C' is classification and type 'R' is regression. The number of features is given after encoding and selection. Since this number may vary between trials, we average it on the 5 trials for the ANOVA selection. Target is the name of the feature to predict in the original database or a formula to build a new feature to predict from the existing ones.}
\begin{tabularx}{\textwidth}{llllllllXX}
\toprule
                      &                   & \rot{Score} & \rot{Scorer} & \rot{Selection} & \rot{Number of samples} & \rot{Number of features} & \rot{Type} &                                                                                                                                                                                                                                                                                                                                                         Target &                                                                                                        Description \\
Database & Task &             &              &                 &                         &                          &            &                                                                                                                                                                                                                                                                                                                                                                &                                                                                                                    \\
\midrule
\rotsmash{Traumabase} & death\_screening &        0.96 &          AUC &               A &                   12341 &                       98 &          C &                                                                                                                                                                                                                                                                                                                                                   Décès == Oui &                                                                                     Predict the death of patients. \\
\bottomrule
\end{tabularx}

    \label{tab:tasks}
\end{table*}

\setlength{\tabcolsep}{3pt}
\begin{table*}
    \caption{\textbf{Scores and ranks of the tree based methods} described in Table~\ref{tab:methods:tree}.}
    \label{tab:tree:scores-ranks}

    \begin{subtable}{1\textwidth}
    \caption{\textbf{Scores} relative to the absolute reference score plotted in Figure~\ref{fig:finding:trees:score}. Values in bold are the reference scores and are absolute. Other scores are given relative to the reference score of their task and size.}
    \label{tab:tree:scores}
    \centering
    \begin{tabularx}{\textwidth}{lllllllllllllll}
\toprule
                          & Database & \multicolumn{5}{l}{Traumabase} & \multicolumn{5}{l}{UKBB} & \multicolumn{2}{l}{MIMIC} &                       NHIS \\
                          & Task &     \rot{death\_screening} &                 \rot{hemo} &      \rot{hemo\_screening} &  \rot{platelet\_screening} &    \rot{septic\_screening} &           \rot{breast\_25} &    \rot{breast\_screening} &     \rot{fluid\_screening} & \rot{parkinson\_screening} &      \rot{skin\_screening} &      \rot{hemo\_screening} &    \rot{septic\_screening} &    \rot{income\_screening} \\
Size & Method &                            &                            &                            &                            &                            &                            &                            &                            &                            &                            &                            &                            &                            \\
\midrule
2500 & MIA &                      -8e-5 &                      +5e-3 &                      -1e-3 &                      -4e-4 &                      +4e-3 &                      +7e-3 &                      +2e-3 &                      +5e-5 &                      +2e-3 &                      -3e-3 &                      +3e-3 &                      +8e-3 &                      +3e-3 \\
                          & Mean &                      -8e-4 &                      +1e-3 &                      -1e-3 &                      -2e-3 &                      -3e-3 &                      +6e-3 &                      -1e-3 &                      -2e-3 &                      +2e-3 &                      -2e-3 &                      -1e-2 &                      -6e-4 &                      -5e-3 \\
\bottomrule
\end{tabularx}

    \end{subtable}
\end{table*}

\setlength{\tabcolsep}{2.4pt}
\begin{table*}
    \ContinuedFloat
    \begin{subtable}{1\textwidth}
    \caption{\textbf{Ranks} computed from the relative scores in Supplementary Table~\ref{tab:tree:scores}. Best average ranks are in bold.}
    \label{tab:tree:ranks}
    \centering
    \begin{tabularx}{\textwidth}{lllllllllllllll@{\hskip 0.15in}llll@{\hskip 0.20in}l}
\toprule
                          & Database & \multicolumn{5}{l}{Traumabase} & \multicolumn{5}{l}{UKBB} & \multicolumn{2}{l}{MIMIC} &                    NHIS & \multicolumn{5}{l}{Average} \\
                          & Task & \rot{death\_screening} &    \rot{hemo} & \rot{hemo\_screening} & \rot{platelet\_screening} & \rot{septic\_screening} & \rot{breast\_25} & \rot{breast\_screening} & \rot{fluid\_screening} & \rot{parkinson\_screening} & \rot{skin\_screening} & \rot{hemo\_screening} & \rot{septic\_screening} & \rot{income\_screening} & \rot{Traumabase} &    \rot{UKBB} &   \rot{MIMIC} &    \rot{NHIS} &     \rot{All} \\
Size & Method &                        &               &                       &                           &                         &                  &                         &                        &                            &                       &                       &                         &                         &                  &               &               &               &               \\
\midrule
2500 & MIA &                      5 &             1 &                     6 &                         4 &                       4 &                1 &                       4 &                      4 &                          2 &                    10 &                     5 &                       5 &                       4 &              4.0 &           4.2 &           5.0 &           4.0 &           4.3 \\
                          & Mean &                      9 &             6 &                     8 &                         7 &                      10 &                5 &                       8 &                      7 &                          3 &                     7 &                     9 &                       8 &                       8 &              8.0 &           6.0 &           8.5 &           8.0 &           7.6 \\
\bottomrule
\end{tabularx}

    \end{subtable}
\end{table*}

\setlength{\tabcolsep}{3.53pt}
\begin{table*}
    \caption{\textbf{Scores and ranks of gradient-boosted trees+MIA compared to linear methods} described in Supplementary Table~\ref{tab:methods:linear}.\\
    We removed one outlier fold from one trial for the methods Linear+Iter and Linear+Iter+mask for the "task platelet\_screening" at size $n$=2\,500. Others are unchanged.}\label{tab:linear:scores-ranks}

    \begin{subtable}{1\textwidth}
    \caption{\textbf{Scores} relative to the absolute reference score and plotted in Supplementary Figure~\ref{fig:finding:linear:score}. Values in bold are the reference scores and are absolute. Other scores are given relative to the reference score of their task and size.}
    \label{tab:linear:scores}
    \centering
\begin{tabularx}{\textwidth}{lllllllllllllll}
\toprule
                          & Database & \multicolumn{5}{l}{Traumabase} & \multicolumn{5}{l}{UKBB} & \multicolumn{2}{l}{MIMIC} &                       NHIS \\
                          & Task &     \rot{death\_screening} &                 \rot{hemo} &      \rot{hemo\_screening} &  \rot{platelet\_screening} &    \rot{septic\_screening} &           \rot{breast\_25} &    \rot{breast\_screening} &     \rot{fluid\_screening} & \rot{parkinson\_screening} &      \rot{skin\_screening} &      \rot{hemo\_screening} &    \rot{septic\_screening} &    \rot{income\_screening} \\
Size & Method &                            &                            &                            &                            &                            &                            &                            &                            &                            &                            &                            &                            &                            \\
\midrule
2500 & Boosted trees+MIA &                      +1e-2 &                      +9e-3 &                      +2e-2 &                      +1e-1 &                      +5e-2 &                      +8e-2 &                      -2e-3 &                      +6e-2 &                      -8e-3 &                      +2e-2 &                      +1e-1 &                      +1e-1 &                      +5e-2 \\
                          & Linear+Mean &                      +8e-4 &                      -3e-3 &                      -3e-3 &                      +8e-3 &                      -4e-3 &                      -6e-3 &                      +2e-4 &                      -1e-2 &                      +2e-3 &                      -6e-3 &                      -1e-2 &                      -2e-2 &                      -2e-2 \\
\bottomrule
\end{tabularx}

    \end{subtable}
\end{table*}

\setlength{\tabcolsep}{4.22pt}
\begin{table*}
    \ContinuedFloat
    \begin{subtable}{1\textwidth}
    \caption{\textbf{Ranks} computed from relative scores in Supplementary Table~\ref{tab:linear:scores}. Best average ranks are in bold.}
    \label{tab:linear:ranks}
    \centering
    \begin{tabularx}{\textwidth}{lllllllllllllll@{\hskip 0.15in}llll@{\hskip 0.20in}l}
\toprule
                          & Database & \multicolumn{5}{l}{Traumabase} & \multicolumn{5}{l}{UKBB} & \multicolumn{2}{l}{MIMIC} &                    NHIS & \multicolumn{5}{l}{Average} \\
                          & Task & \rot{death\_screening} &    \rot{hemo} & \rot{hemo\_screening} & \rot{platelet\_screening} & \rot{septic\_screening} & \rot{breast\_25} & \rot{breast\_screening} & \rot{fluid\_screening} & \rot{parkinson\_screening} & \rot{skin\_screening} & \rot{hemo\_screening} & \rot{septic\_screening} & \rot{income\_screening} & \rot{Traumabase} &    \rot{UKBB} &   \rot{MIMIC} &    \rot{NHIS} &     \rot{All} \\
Size & Method &                        &               &                       &                           &                         &                  &                         &                        &                            &                       &                       &                         &                         &                  &               &               &               &               \\
\midrule
2500 & Boosted trees+MIA &                      1 &             1 &                     1 &                         1 &                       1 &                1 &                       7 &                      1 &                          7 &                     1 &                     1 &                       1 &                       1 &     \textbf{1.0} &  \textbf{3.4} &  \textbf{1.0} &  \textbf{1.0} &  \textbf{1.6} \\
                          & Linear+Mean &                      4 &             7 &                     6 &                         3 &                       6 &                5 &                       6 &                      7 &                          3 &                     7 &                     3 &                       6 &                       6 &              5.2 &           5.6 &           4.5 &           6.0 &           5.3 \\
\bottomrule
\end{tabularx}

    \end{subtable}
\end{table*}

\end{document}